\pdfoutput=1

\documentclass[11pt]{article}

\usepackage[]{acl}

\usepackage{times}
\usepackage{latexsym}

\usepackage[T1]{fontenc}

\usepackage[utf8]{inputenc}
\usepackage{graphicx}
\usepackage{adjustbox}
\usepackage{booktabs}
\usepackage{array, makecell}
\usepackage{boldline}
\usepackage{subcaption}
\usepackage{amsmath}

\usepackage{microtype}

\title{NER-BERT: A Pre-trained Model for Low-Resource Entity Tagging}

\author{Zihan Liu$^1$, Feijun Jiang$^2$, Yuxiang Hu$^2$, Chen Shi$^2$, Pascale Fung$^1$ \\
$^1$The Hong Kong University of Science and Technology\\
$^2$Alibaba Group \\
\texttt{zihan.liu@connect.ust.hk}
}

\begin{document}
\maketitle
\begin{abstract}
Named entity recognition (NER) models generally perform poorly when large training datasets are unavailable for low-resource domains. Recently, pre-training a large-scale language model has become a promising direction for coping with the data scarcity issue. However, the underlying discrepancies between the language modeling and NER task could limit the models' performance, and pre-training for the NER task has rarely been studied since the collected NER datasets are generally small or large but with low quality. In this paper, we construct a massive NER corpus with a relatively high quality, and we pre-train a NER-BERT model based on the created dataset. Experimental results show that our pre-trained model can significantly outperform BERT~\cite{devlin2019bert} as well as other strong baselines in low-resource scenarios across nine diverse domains. Moreover, a visualization of entity representations further indicates the effectiveness of NER-BERT for categorizing a variety of entities.

\end{abstract}

\section{Introduction}
Named entity recognition\footnote{This term is interchangeable with ``entity tagging''.} (NER) plays an important role in information extraction and text processing. Current NER systems heavily rely on large training datasets to achieve good performance~\cite{lample2016neural,chiu2016named,ma2016end,yadav2018survey,li2020unified}, and a well-designed NER model normally has a poor generalization ability on low-resource domains, where large numbers of training data are unavailable~\cite{jia2019cross,liu2021crossner}. Given that collecting numerous NER training data is not just expensive but also time-consuming, it is essential to construct a NER model that can quickly adapt to low-resource domains using only a few data examples.

Recently, pre-training a large-scale language model~\cite{devlin2019bert,liu2019roberta} has been shown to be effective in a data scarcity scenario~\cite{ma2019domain,radford2019language,chen2020few}.
However, the underlying discrepancies between the language modeling and the NER task could limit the performance of pre-trained language models on this task. Unfortunately, conducting a NER-specific pre-training has rarely been studied because constructing a large-scale and high-quality corpus for this purpose is not a simple task.

Although there are plenty of publicly available NER datasets, they generally have different annotation schemes and different entity categories.
For example, the CoNLL2003 dataset~\cite{sang2003introduction} has the ``miscellaneous'' entity category, which the Broad Twitter dataset~\cite{derczynski2016broad} lacks, and the WNUT2017 dataset~\cite{derczynski2017results} has ``corporation'' and ``group'' entity categories, while many other datasets~\cite{sang2003introduction,lu2018visual} use the ``organization'' entity type.
Thus, it is difficult to unify the annotation scheme for all datasets, and jointly training models on different schemes will confuse the model in categorizing entities.
In addition, the existing NER datasets are much smaller than those of the plain text used for the language modeling task, which will result in a less effective pre-training.

Instead of utilizing manually annotated NER datasets, a few previous studies~\cite{cao2019low,mengge2020coarse} have focused on leveraging weakly-labeled NER data constructed from Wikipedia to enhance the model's performance. 
\citet{cao2019low} generated the weakly-labeled data based on Wikipedia anchors and a taxonomy, but the quality of the produced data is relatively low and the number of entity categories is limited. To cope with these issues, \citet{mengge2020coarse} leveraged a gazetteer to obtain coarse-grained entities and k-means clustering to further mine the fine-grained entities. However, obtaining fine-grained labels based on clustering algorithms is not stable, which could limit the effectiveness of pre-training.

In this work, we first aim to construct a large-scale NER dataset with a relatively high quality and abundant entity categories. After that, our goal is to prove that using the created dataset to pre-train an entity tagging model can outperform pre-trained language models on the low-resource NER task.
Similar to~\citet{cao2019low}, we build the NER dataset based on the Wikipedia corpus. 
To improve the quality and increase the number of entity categories, we utilize the DBpedia Ontology~\cite{mendes2012dbpedia} to assist in categorizing entities in the Wikipedia corpus.
Eventually, we obtain around 16 million NER training examples, and then we continue pre-training BERT on the NER task using the constructed data to build NER-BERT.

We emphasize that the focus of this paper is not to achieve state-of-the-art results, but to show the effectiveness of entity tagging-based pre-training using our constructed corpus, since current state-of-the-art NER models are constructed on top of pre-trained language models (e.g., BERT~\cite{devlin2019bert}) which can be easily replaced by our NER-BERT. Therefore, we simply add a linear layer instead of many complex components on top of the pre-trained models when fine-tuning them on the downstream NER task.
We evaluate our model and baselines on nine diverse domains (e.g., literature, biomedical, and Twitter) of the NER task and show that our model can surpass BERT and other strong baselines such as cross-domain language modeling~\cite{jia2019cross} and domain-adaptive pre-training~\cite{gururangan2020don,liu2021crossner}. Furthermore, we conduct extensive experiments in terms of different low-resource levels across multiple diverse target domains and demonstrate that NER-BERT has a powerful few-shot adaptation ability to target domains when only a few training data are available.
Additionally, we visualize the entity representations for NER-BERT and baselines to further prove the effectiveness of our NER pre-training.
Moreover, we will release our constructed dataset and pre-trained model to facilitate future research in this area.

\section{Corpus Construction}
In this section, we introduce how we construct a massive NER dataset with a relatively high quality for the model pre-training. We first discuss the limitations of using existing NER datasets for pre-training, as well as the data sources we use for the NER corpus construction. Then, we detail the process of entity categorization and data filtering. Finally, we describe how we balance the data for different entity categories.

\subsection{Data Sources for Pre-training}
To create a massive NER corpus, a straightforward idea is to integrate the multiple existing human-annotated NER datasets. However, we find several limitations of using them for pre-training.
First, these datasets are much smaller (generally around or less than 10K data examples) than the datasets used for pre-training models (e.g., BERT is trained on BookCorpus (800M words) and Wikipedia (2500M words)).
Second, these datasets generally have different annotation schemes. Some entity categories in certain datasets do not exist in the other datasets, and different entity categories across datasets could have overlaps. For example, the ``corporation'' and ``group'' entity categories in WNUT2017~\cite{derczynski2017results} have overlaps with the ``organization'' entity in many other datasets~\cite{sang2003introduction,lu2018visual}.
It is difficult to unify the annotation schemes for all datasets and jointly training models on them will easily confuse them in categorizing entities.
Third, the entity categories in these datasets are generally limited to only three or four types (e.g., location, person and organization), which makes the pre-training on them imperfect since the model cannot learn the information of various entity types.

To mitigate these limitations, we aim to find a large-scale data source that contains abundant entity categories and use a unified scheme to extract the entities from it so as to ensure the constructed dataset is reliable for pre-training.
Wikipedia naturally contains plentiful entity information since we can easily find entities by looking for consecutive words that have hyperlinks (i.e., anchors) on them.
Since anchors do not provide the information of concrete entity types, previous works leverage a taxonomy~\cite{cao2019low} or gazetteer~\cite{mengge2020coarse} to categorize the entities. However, the total entity categories are still limited to only a few types (e.g., person, location, and organization). To enlarge the number of entity categories, we propose to leverage the DBpedia Ontology~\cite{mendes2012dbpedia} to help categorize the entities. We choose the DBpedia Ontology because it contains 320 entity types extracted from Wikipedia, and there are 3.64 million entities categorized based on these entity types, which ensures a large coverage of Wikipedia entities and a relatively high quality of the constructed NER corpus.

\begin{figure*}[!ht]
    \centering
    \resizebox{0.999\textwidth}{!}{  \includegraphics{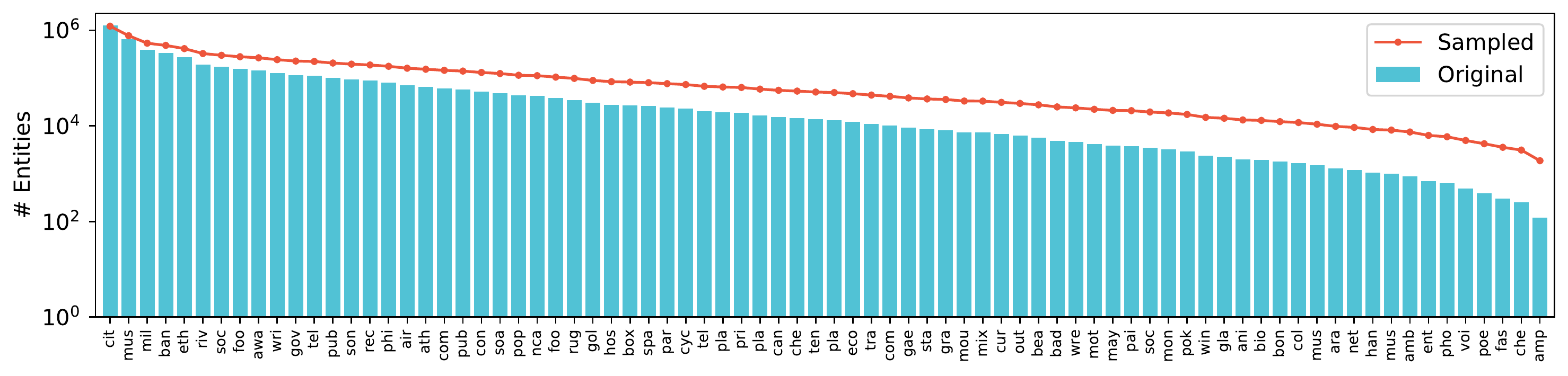}
    }
    \caption{Number of entities before and after the sampling. X-axis is a list of categories that are denoted by their first three characters. Note that we only show an evenly spaced quarter of all categories due to the length limit. The whole table along with the category list with full category names can be found in the Appendix~\ref{Appendix_D}.}
    \label{fig:category}
\end{figure*}

\subsection{Entity Categorization}
We tokenize the Wikipedia articles into sentences by using the sentence tokenization in NLTK~\cite{loper2002nltk}, and then we combine the Wikipedia anchors and the DBpedia Ontology to conduct the entity categorization for each sentence.
Concretely, when we find consecutive words (or a single word) with an anchor, we consider them (or it) as an entity and check whether this entity exists in the DBpedia Ontology. If so, we will categorize this entity with the corresponding entity type. Otherwise, we will give it a special entity label (\texttt{ENTITY}) to denote that it is an entity. 
Note that we only use the DBpedia Ontology to categorize words with anchors, instead of all words in Wikipedia, to ensure the quality of the categorization. Consecutive words with anchors are highly likely to be entities, and if they can be found in the DBpedia Ontology, it ensures the correctness of the categorization.
In addition, categorizing for all words is not just time-consuming, but will also misclassify words that are accidentally matched in the DBpedia Ontology due to having the same spellings but are actually not entities. 

\begin{table}[]
\renewcommand{\arraystretch}{1.03}
\centering
\begin{adjustbox}{width={0.49\textwidth},totalheight={\textheight},keepaspectratio}
\begin{tabular}{cccc}
\toprule
\# Examples & \# Tokens & \# Categories  & Corpus Size \\
\cmidrule(lr){1-1}\cmidrule(lr){2-2}\cmidrule(lr){3-3}\cmidrule(lr){4-4}
16.3M        & 475.6M     & 315   & 4.2GB   \\
\bottomrule
\end{tabular}
\end{adjustbox}
\caption{Data statistics for the collected corpus.}
\label{tab:statistics}
\end{table}

\subsection{Data Filtering}
We further conduct several data filtering processes to improve the data quality for pre-training.

First, we discard a few scarce entity categories for which very few (less than ten) corresponding entities can be found in Wikipedia. This is because it is difficult for the model to capture the features of these categories due to the data scarcity issue, and data examples containing these categories will become noisy examples, which could slightly hurt the effectiveness of the pre-training. 

Second, we filter sentences that do not have any entities or only have special entity labels (\texttt{ENTITY}). Considering that large numbers of words are being labeled as \texttt{ENTITY} since they do not exist in the DBpedia Ontology, we want to increase the ratio of data examples that contain concrete entity categories in order to encourage the model to learn the knowledge from diverse entity categories.

Third, for the same purpose as the second method, we use a certain probability to filter sentences where all the entities frequently exist in the corpus and there are too many \texttt{ENTITY} labels simultaneously. Concretely, the probabilities to filter sentences are illustrated as follows:
\begin{align*}
    & \texttt{prob} = 0.3 \hspace{5mm} \texttt{if all \& num = 3}, \\
    & \texttt{prob} = 0.5 \hspace{5mm} \texttt{if all \& num = 4}, \\
    & \texttt{prob} = 0.7 \hspace{5mm} \texttt{if all \& num > 4},
\end{align*}
where \texttt{all} means that all the entities in a sentence are within the top 20 frequent entity categories (including \texttt{ENTITY}), and \texttt{num} denotes the number of \texttt{ENTITY} in a sentence. The data statistics after this data filtering process are illustrated in Table~\ref{tab:statistics}.

\subsection{Data Balancing}
After the data filtering, there still exists a data imbalance issue across the entity categories. Pre-training on imbalanced data could cause model bias and lead to a less effective pre-training.
To alleviate this issue, we follow~\citet{lample2019cross},\citet{conneau2020unsupervised}, and \citet{xue2020mt5} to boost the entity numbers of low-resource categories by sampling examples according to probability $p(E) \propto |E|^\alpha $, where $p(E)$ is the probability of sampling sentences that contain category $E$, $|E|$ is the number of entities in the category, and we set $\alpha=0.7$ in the sampling process. As illustrated in Figure~\ref{fig:category}, we can observe that the data imbalance issue for entity categories is alleviated after the sampling process.

\section{NER-BERT}
In this section, we describe how we build the NER-BERT model based on the collected NER corpus, and how we fine-tune NER-BERT to downstream NER tasks.

\subsection{Pre-training}
We pre-train NER-BERT based on the architecture of BERT-Base-Cased~\cite{devlin2019bert} and we replace its language model head with an entity tagging head that covers all the entity categories in the constructed corpus.\footnote{We place the concrete entity categories in Appendix~\ref{Appendix_B}.}
To leverage the powerful language understanding ability of BERT, we initialize NER-BERT with the pre-trained weight from BERT (while the entity tagging head has to be pre-trained from scratch). Unlike the pre-training of BERT, which randomly masks some tokens in the input sequence and then trains the model to predict the masked tokens, we directly train NER-BERT to conduct the sequence labeling task to detect and categorize entities using the constructed NER corpus. 
Note that the NER corpus we build can be used to conduct pre-training on the architecture of any existing language model, and we select BERT simply because it is the most widely used pre-trained model in the natural language processing research and many task-specific pre-trained models are built based on it~\cite{beltagy2019scibert,sun2019videobert,Su2020VLBERT,wu2020tod}.

\subsection{Fine-tuning}
In the fine-tuning stage, we first replace the entity tagging head of NER-BERT with a randomly initialized new head that covers all the entity categories of the target domain in the downstream NER task. Then, we directly fine-tune the model using the target domain's training data.

\begin{table*}[!th]
\renewcommand{\arraystretch}{1.3}
\centering
\begin{adjustbox}{width={0.99\textwidth},totalheight={\textheight},keepaspectratio}
\begin{tabular}{lcccccccccccc}
\toprule
\multicolumn{1}{l|}{\textbf{Models}}     & \textbf{Pol.}  & \textbf{Sci.}  & \textbf{Mus.}                          & \textbf{Lit.}                          & \textbf{AI}                            & \textbf{Twi.}                          & \textbf{BTwi.}                         & \textbf{BioCG}                         & \textbf{BioPC}                         & \textbf{Fin.}                          & \multicolumn{1}{c|}{\textbf{Def.}}  & \textbf{Avg.}  \\ \midrule \midrule
\multicolumn{13}{c}{\textit{\textbf{Directly Fine-tune on Target Domains (Target Only)}}}     \\ \cmidrule[0.12ex]{1-13}
\multicolumn{1}{l|}{BERT}        & 66.56 & 63.73 & 66.59 & 59.95 & 50.37 & 83.34 & 75.61 & 78.05 & 83.20 & 76.23                         & \multicolumn{1}{c|}{68.42} & 70.19 \\
\multicolumn{1}{l|}{DAPT$^\dagger$}        & 70.45 & 67.59 & 73.39                         & 64.96                         & 56.36                         & -                             & -                             & -                             & -                             & -                             & \multicolumn{1}{c|}{-}     & -     \\
\multicolumn{1}{l|}{NER-BERT$_{\text{4types}}$}   & 70.76 & 68.31 & 71.02                         & 63.91                         & 57.03                         & 83.40 & 77.26 & 78.80 & 83.91 & 77.55 & \multicolumn{1}{c|}{69.22} & 72.83 \\
\multicolumn{1}{l|}{NER-BERT$_{\text{212types}}$} & \textbf{73.81} & 71.09 & 75.98                         & \textbf{68.13}                         & 58.58                         & 83.46                         & 77.06                         & 79.20                         & 85.11                         & 77.85                         & \multicolumn{1}{c|}{70.28} & 74.60 \\ \cmidrule(){1-13}
\multicolumn{1}{l|}{NER-BERT}    & 73.69 & \textbf{71.90} & \textbf{76.23}                         & 67.85                         & \textbf{60.39}                         & \textbf{83.59}                         & \textbf{77.30}                         & \textbf{79.86}                         & \textbf{85.35}                         & \textbf{78.72}                         & \multicolumn{1}{c|}{\textbf{70.79}} & \textbf{75.06} \\ \midrule \midrule
\multicolumn{13}{c}{\textit{\textbf{Train on the Source Domain then Fine-tune on Target Domains (Source \& Target)}}}      \\ \cmidrule[0.12ex]{1-13}
\multicolumn{1}{l|}{BERT}        & 68.71 & 64.94 & 68.30                         & 63.63                         & 58.88                         & 83.77                         & 77.28                         & 78.59                         & 84.02                         & 75.97                         & \multicolumn{1}{c|}{69.57} & 72.15 \\
\multicolumn{1}{l|}{CDLM$^\ddagger$}        & 68.44 & 64.31 & 63.56                         & 59.59                         & 53.70                         & -                             & -                             & 79.86                         & 85.54                         & -                             & \multicolumn{1}{c|}{-}     & -     \\
\multicolumn{1}{l|}{DAPT$^\dagger$}        & 72.05 & 68.78 & 75.71                         & 69.04                         & 62.56                         & -                             & -                             & -                             & -                             & -                             & \multicolumn{1}{c|}{-}     & -     \\
\multicolumn{1}{l|}{NER-BERT$_{\text{4types}}$}   & 71.98 & 69.27 & 75.46                         & 66.37                         & 59.03                         & 83.85    & 77.70    & 79.57    & 84.29    & 77.88                         & \multicolumn{1}{c|}{70.21} & 74.14 \\
\multicolumn{1}{l|}{NER-BERT$_{\text{212types}}$} & 75.09 & \textbf{72.13} & 79.49                         & 71.33                         & 63.27                         & 83.77                         & 77.74                         & 79.63                         & 84.97                         & 78.30                         & \multicolumn{1}{c|}{70.70}  & 76.04 \\ \cmidrule(){1-13}
\multicolumn{1}{l|}{NER-BERT}    & \textbf{76.12} & 72.10 & \textbf{80.20}                         & \textbf{71.90}                         & \textbf{63.34}                         & \textbf{83.97}                         & \textbf{77.76}                         & \textbf{80.16}                         & \textbf{85.86}                         & \textbf{78.39}                         & \multicolumn{1}{c|}{\textbf{71.59}} & \textbf{76.49} \\ 
\bottomrule
\end{tabular}
\end{adjustbox}
\caption{F1-scores on eleven domains (containing nine diverse domains). Note that we use the first three characters to denote most of the domains. ``BTwi.'' denotes Broad Twitter, and ``BioCG'' and ``BioPC'' denote BioNLP13CG and BioNLP13PC, respectively. $^\dagger$ Results are taken from~\citet{liu2021crossner}. $^\ddagger$ Results for the CrossNER dataset are taken from~\citet{liu2021crossner}, and those for BioCG and BioPC are taken from~\citet{jia2019cross}.}
\label{tab:main_results}
\end{table*}

\section{Experiments}

\subsection{Datasets \& Domains}
We conduct experiments on the CrossNER~\cite{liu2021crossner}, Twitter~\cite{lu2018visual}, Broad Twitter~\cite{derczynski2016broad}, BioNLP13PC and BioNLP13CG~\cite{nedellec2013overview}, SEC-filings~\cite{alvarado2015domain}, and re3d\footnote{\url{https://github.com/dstl/re3d}} datasets, which cover nine diverse domains, as follows:\footnote{The data statistics for these domains are placed in the Appendix~\ref{Appendix_A}.}

\paragraph{Politics (from CrossNER)}
This domain contains ``politician'', ``person'', ``organization'', ``political party'', ``event'', ``election'', ``country'', ``location'', and ``miscellaneous'' entity categories.

\paragraph{Science (from CrossNER)} This domain contains ``scientist'', ``person'', ``university'', ``organization'', ``country'', ``location'', ``discipline'', ``enzyme'', ``protein'', ``chemical compound'', ``chemical element'', ``event'', ``astronomical object'', ``academic journal'', ``award'', ``theory'', and ``miscellaneous'' entity categories.

\paragraph{Music (from CrossNER)} This domain contains ``music genre'', ``song'', ``band'', ``album'', ``artist'', ``instrument'', ``award'', ``event'', ``country'', ``location'', ``organization'', ``person'', and ``miscellaneous'' entity categories.

\paragraph{Literature (from CrossNER)} This domain contains ``book'', ``writer'', ``award'', ``poem'', ``event'', ``magazine'', ``person'', ``location'', ``organization'', ``country'', and ``miscellaneous'' entity categories.

\paragraph{AI (from CrossNER)} This is the artificial intelligence domain which contains ``field'', ``task'', ``product'', ``algorithm'', ``researcher'', ``metrics'', ``university'', ``country'', ``person'', ``organization'', ``location'', and ``miscellaneous'' entity categories.

\paragraph{Twitter}
Both the Twitter and Broad Twitter datasets belong to this domain. Twitter contains ``person'', ``location'', ``organization'', and ``miscellaneous'' categories, while Broad Twitter contains ``person'', ``location'', and ``organization'' categories.

\paragraph{Biomedical} This domain contains the BioNLP13PC and BioNLP13CG datasets. BioNLP13PC mainly consists of five entity types: ``simple chemical'', ``cellular component'', ``gene and gene product'', ``species'' and ``cell''. And BioNLP13CG mainly consists of three entity types: ``simple chemical'', ``cellular component'', and ``gene and gene product''.

\paragraph{Finance (from SEC-filings)}
This domain contains ``person'', ``location'', ``organization'', and ``miscellaneous'' entity categories. 

\paragraph{Defense (from re3d)}
This domain is related to defence and security analysis, and contains ``person'', ``location'', ``organisation'', ``temporal'', ``nationality'', ``documentreference'', ``money'', ``militaryplatform'', ``weapon'', and ``quantity'' entity categories.

\subsection{Experimental Setup}
We consider two experimental settings. First, we directly fine-tune the models to the target domain. Second, we follow~\citet{jia2019cross} and \citet{liu2021crossner} to leverage English CoNLL-2003~\cite{sang2003introduction} as the source domain to further boost the model's performance on target domains. Specifically, we first train the model on this source domain, and then fine-tune it to the target domain. 
In addition, we further study the effectiveness of NER-BERT in the low-resource scenario by conducting few-shot experiments across the Twitter, Biomedical, Finance and Defense domains.

\begin{figure*}[!ht]
\begin{subfigure}{.33\textwidth}
  \centering
  \includegraphics[width=\linewidth]{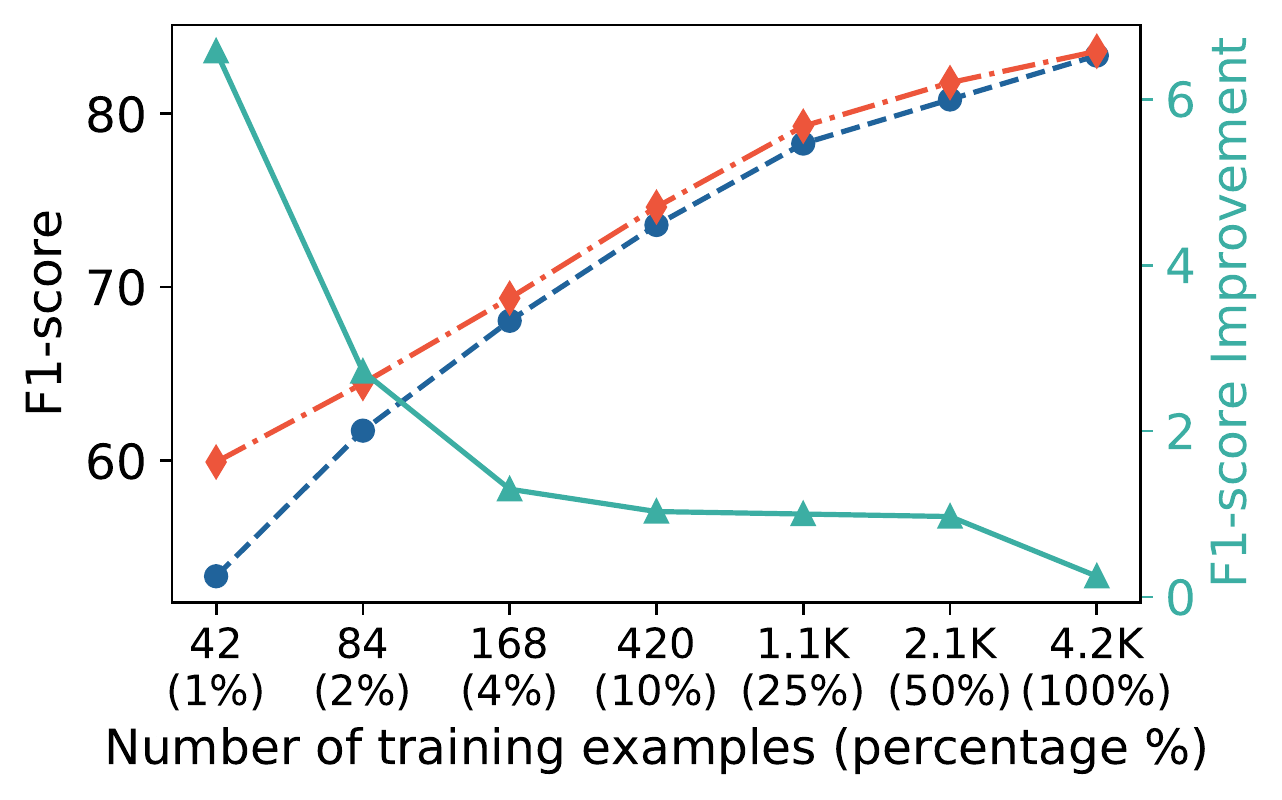}  
  \caption{Twitter}
\end{subfigure}
\begin{subfigure}{.33\textwidth}
  \centering
  \includegraphics[width=\linewidth]{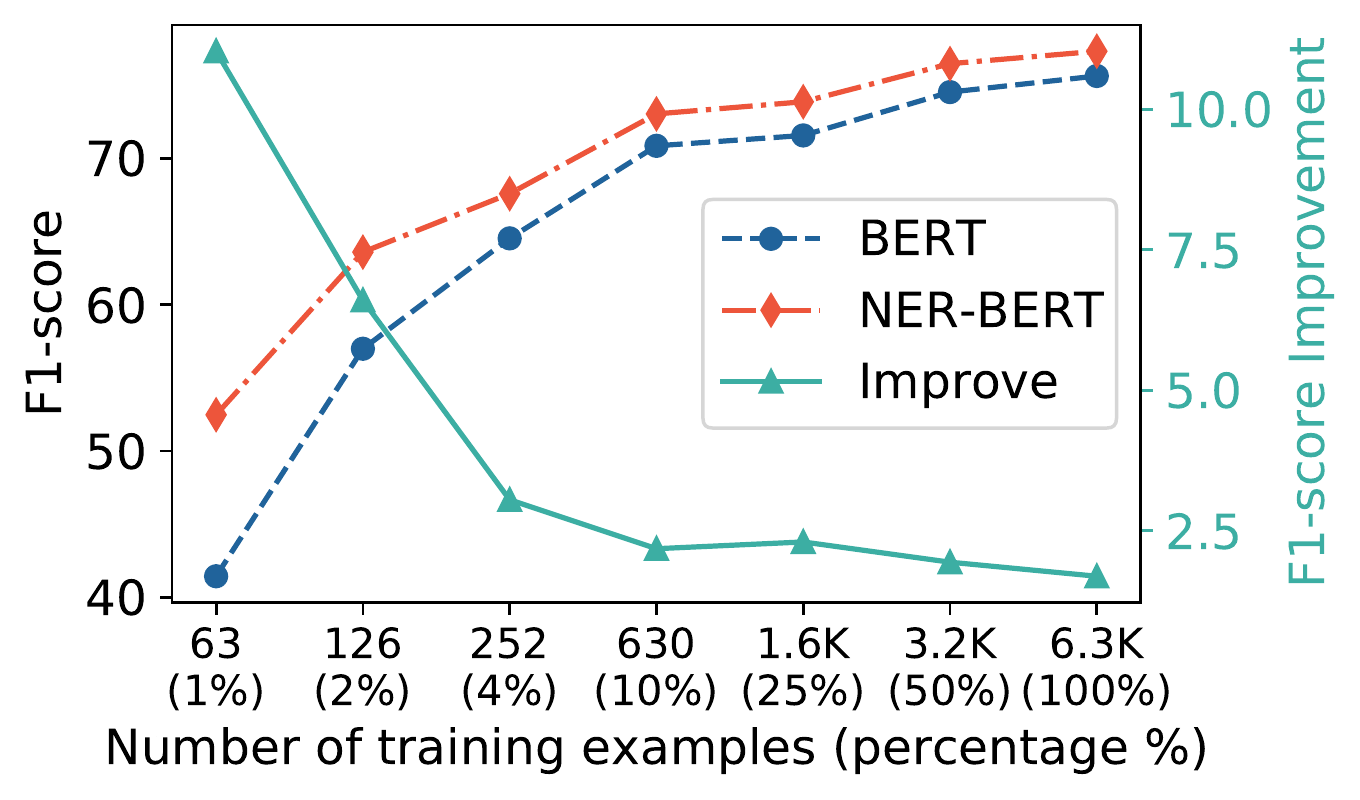}
  \caption{Broad Twitter}
\end{subfigure}
\begin{subfigure}{.33\textwidth}
  \centering
  \includegraphics[width=\linewidth]{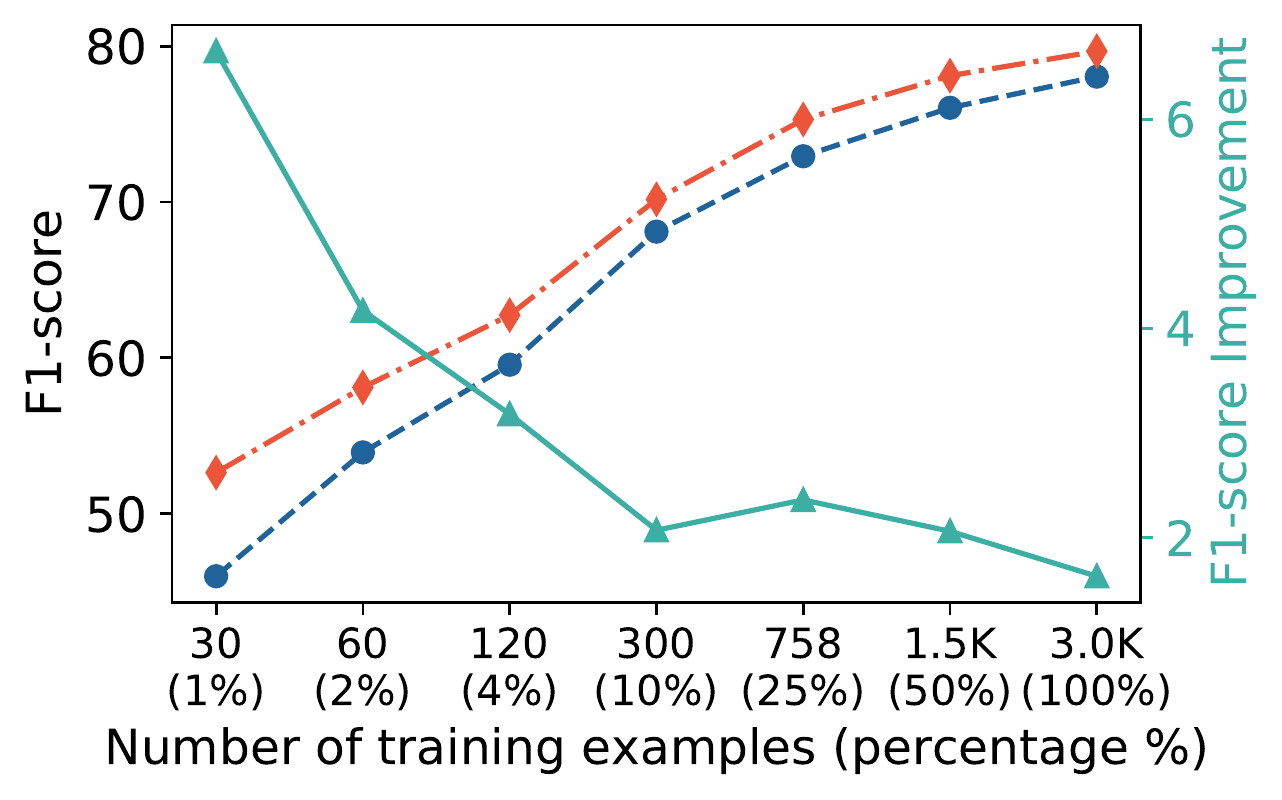}
  \caption{BioNLP13CG}
\end{subfigure}
\begin{subfigure}{.33\textwidth}
  \centering
  \includegraphics[width=\linewidth]{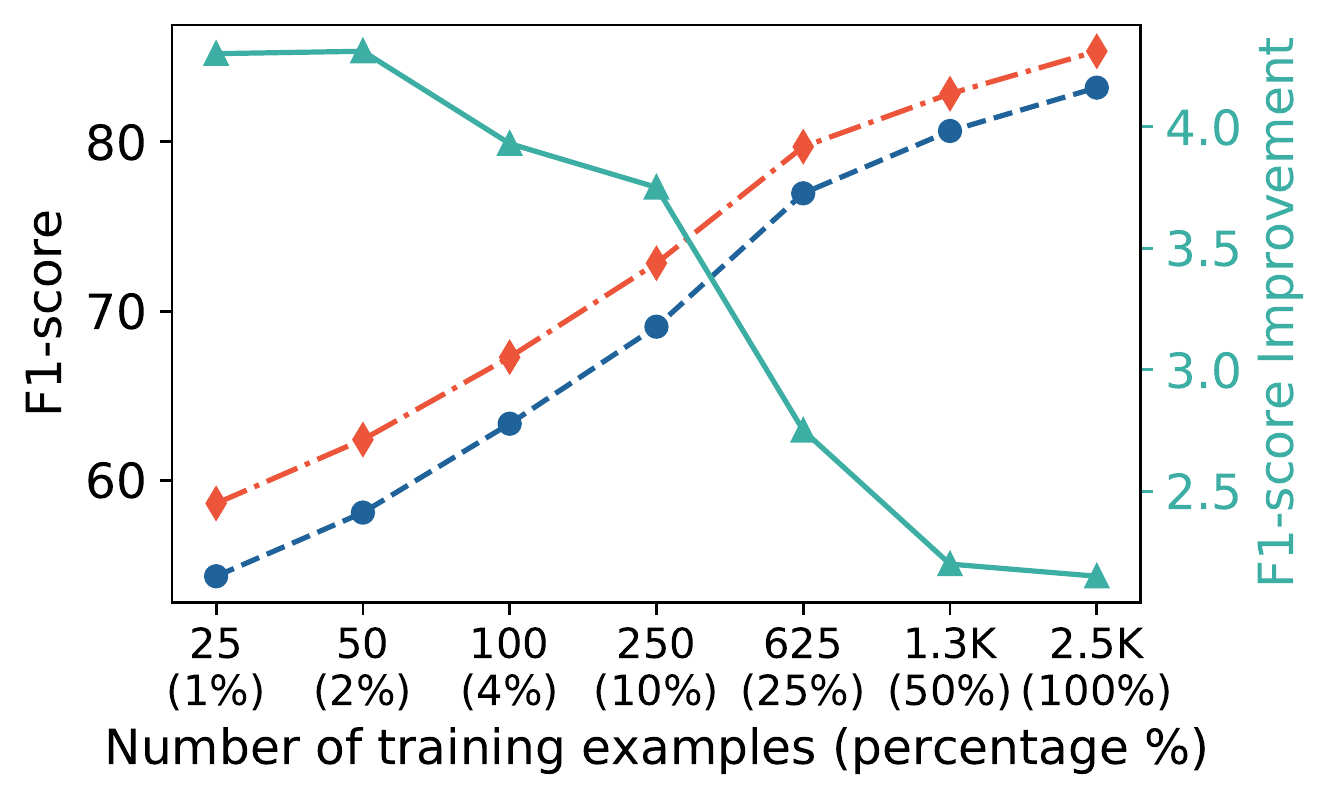}
  \caption{BioNLP13PC}
\end{subfigure}
\begin{subfigure}{.33\textwidth}
  \centering
  \includegraphics[width=\linewidth]{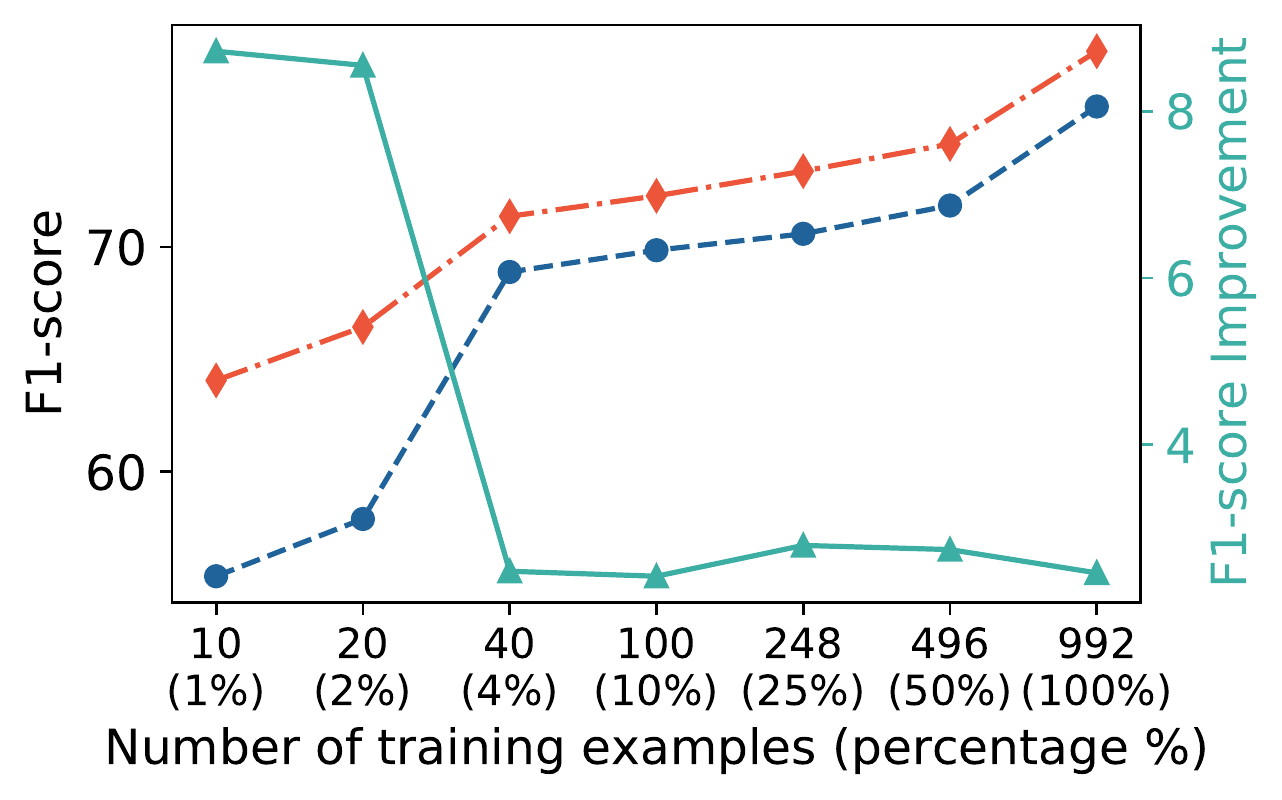}
  \caption{Finance}
\end{subfigure}
\begin{subfigure}{.33\textwidth}
  \centering
  \includegraphics[width=\linewidth]{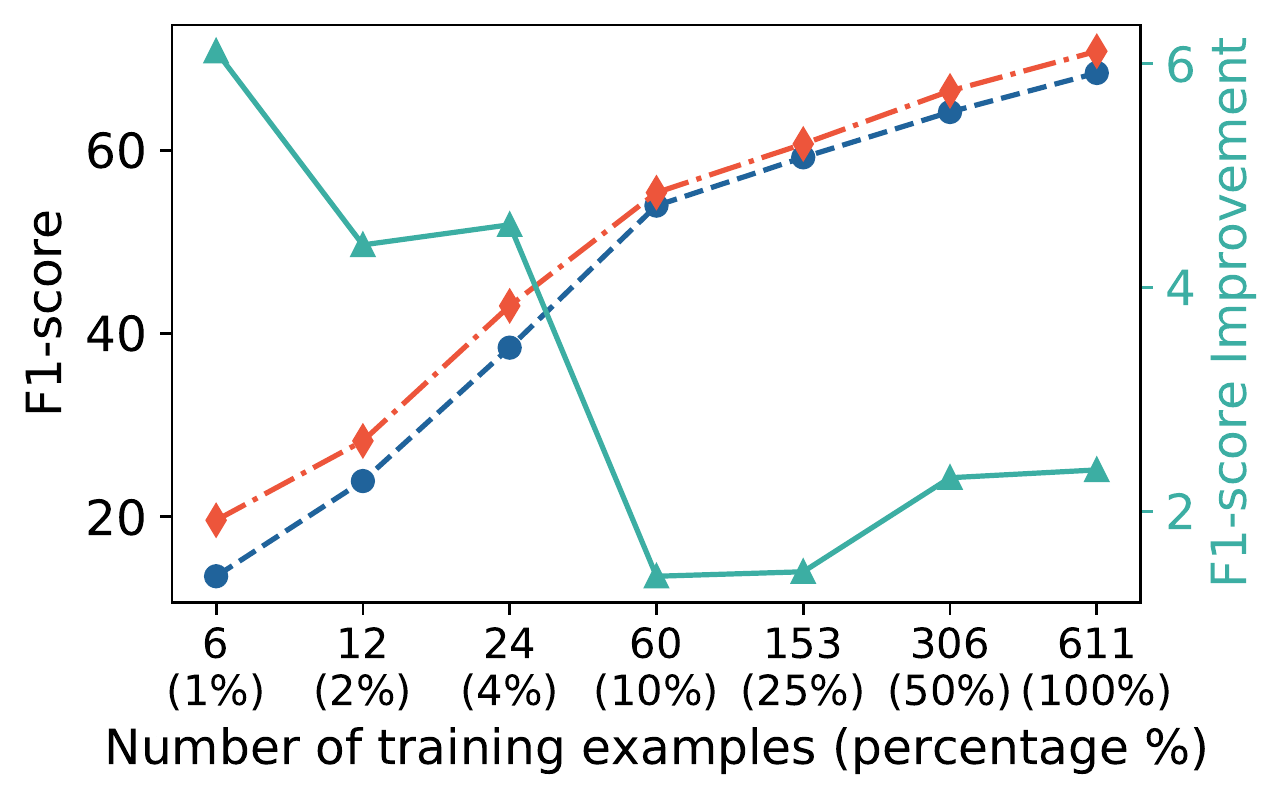}
  \caption{Defense}
\end{subfigure}
\caption{Few-shot F1-scores for BERT and NER-BERT models (shown in left y-axis), and the improvements of NER-BERT over BERT (shown in right y-axis) on a variety of domains.}
\label{fig:few-shot}
\end{figure*}

\subsection{Baselines}
\paragraph{BERT}
As the backbone of our NER-BERT model, BERT~\cite{devlin2019bert} is shown to possess a strong language understanding ability across eleven tasks, including NER. Here, we use the BERT-Base-Cased version since NER-BERT is pre-trained based on it.

\paragraph{Cross-Domain Language Modeling (CDLM)}
This baseline~\cite{jia2019cross} leverages the unlabeled plain text and NER data from both source and target domains to perform cross-domain and cross-task knowledge transfer, which is shown to be effective for NER domain adaptation.

\paragraph{Domain-Adaptive Pre-training (DAPT)}
\citet{liu2021crossner} conducted DAPT based on BERT using a large unlabeled domain-related corpus and evaluated it on the CrossNER dataset. They showed that DAPT can greatly improve the domain adaptation performance based upon BERT.

\paragraph{NER-BERT$_{\text{4types}}$}
We want to compare the effectiveness of using coarse-grained entity types and fine-grained entity types for pre-training. We compressed the entity categories into four general types: ``person'', ``location'', ``organization'', and ``miscellaneous''. To ensure a fair comparison, this model is trained using the same amount of data as NER-BERT. We use the \texttt{ENTITY} category to represent the miscellaneous type, and we classify all the categories into these four types. If a category does not belong to ``person'', ``location'' or ``organization'', we classify it as the \texttt{ENTITY} type. For example, the ``body of water'' category belongs to the ``location'' type, while the ``award'' category will be classified as the \texttt{ENTITY} type.

\paragraph{NER-BERT$_{\text{212types}}$}
Instead of greatly compressing the categories into four types, we only consider merging the most fine-grained categories to lower the number of entity categories. For example, we merge ``tennis tournament'', ``soccer tournament'', ``golf tournament'', etc. into ``sports tournament''. At the end, we compress the number of entity types from 315 to 212, and then use this new category list to pre-train NER-BERT$_{\text{212types}}$.

\subsection{Training Details}
We follow~\citet{devlin2019bert} to conduct the NER task by adding a linear layer on top of pre-trained models to predict entity types. If there exist multiple subwords for a token, we take the representation of its first subword token to make the prediction. We use the Adam optimizer~\cite{kingma2015adam} with a learning rate of 5e-5 for both pre-training and fine-tuning. We use a batch size of 960 for pre-training and a batch size of 32 or 16 for fine-tuning.
We use the BIO label structure, and the F1-score (based on BIO) is used as the evaluation metrics. The pre-training data are randomly divided into a 90:10 percent split for training and validation, and we select the NER-BERT checkpoint that has the best performance on the validation set of the pre-training data. To ensure a fair comparison, all the results for BERT, NER-BERT$_{\text{4types}}$, NER-BERT$_{\text{212types}}$, and NER-BERT are averaged over five runs with the same five random seeds.

\begin{figure*}[!ht]
\centering
\begin{subfigure}{.49\textwidth}
  \centering
  \includegraphics[width=0.9\linewidth]{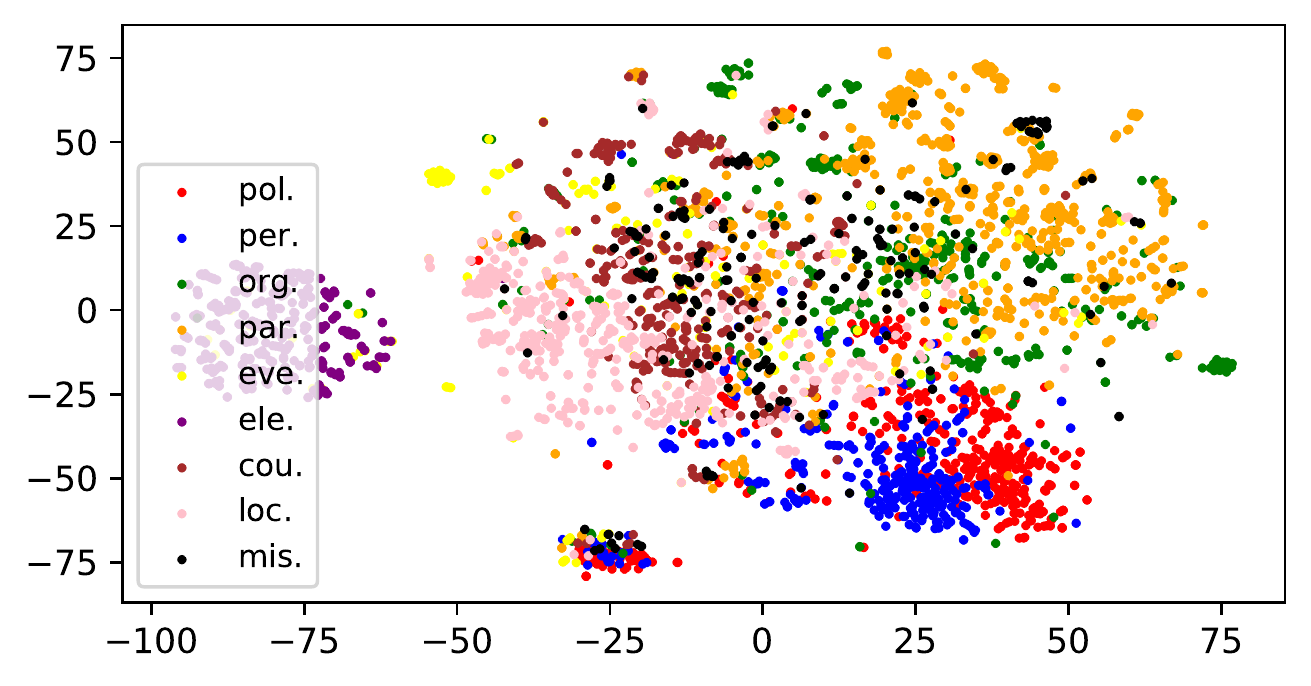}  
  \caption{BERT}
\end{subfigure}
\begin{subfigure}{.49\textwidth}
  \centering
  \includegraphics[width=0.85\linewidth]{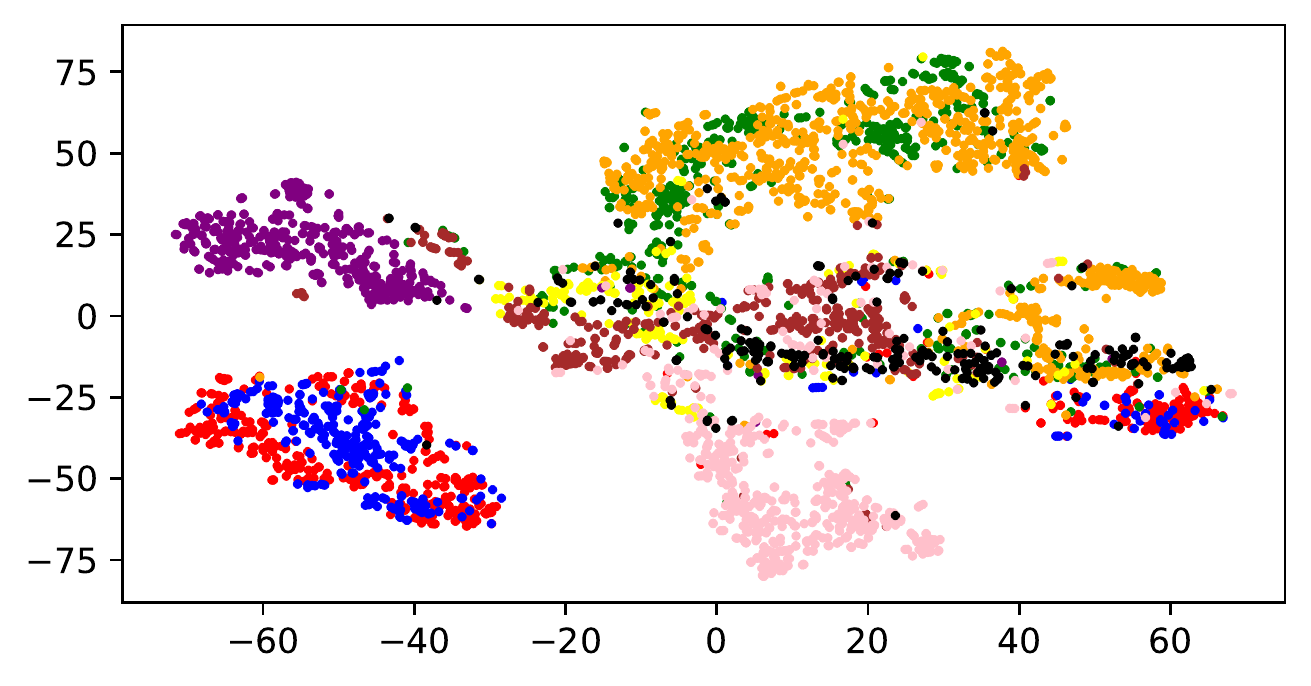}
  \caption{NER-BERT$_{\text{4types}}$}
\end{subfigure}
\begin{subfigure}{.49\textwidth}
  \centering
  \includegraphics[width=0.9\linewidth]{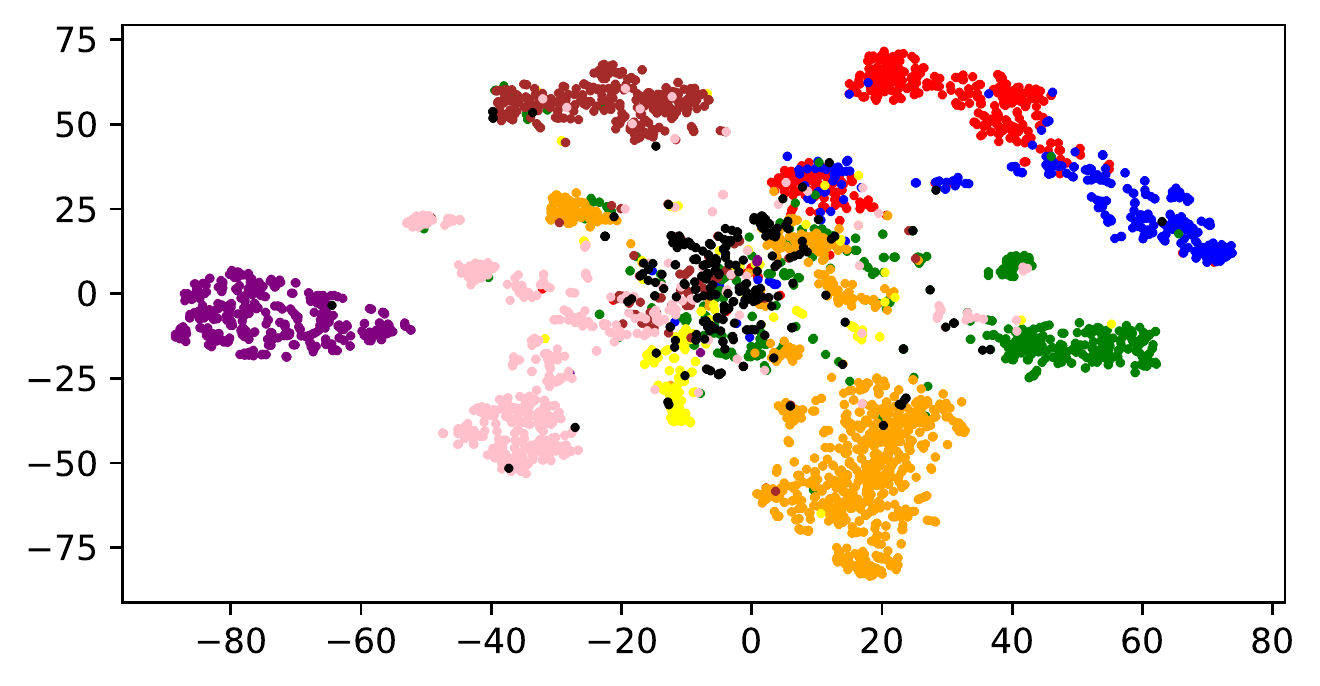}
  \caption{NER-BERT$_{\text{212types}}$}
\end{subfigure}
\begin{subfigure}{.49\textwidth}
  \centering
  \includegraphics[width=0.85\linewidth]{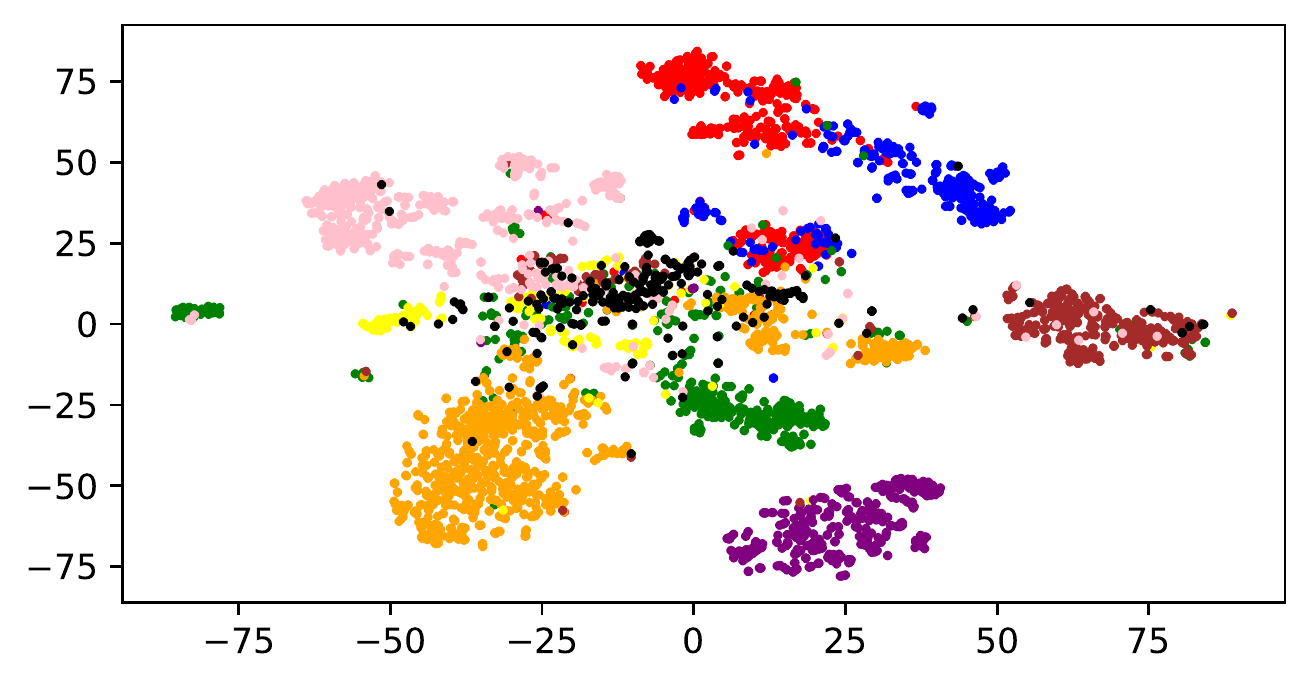}
  \caption{NER-BERT}
\end{subfigure}
\caption{The tSNE visualization of entity representations in the test set of the politics domain.
Different colors denote different entity categories (the legend shows the first three characters of each category).}
\label{fig:visualization}
\end{figure*}

\section{Results \& Discussion}

\subsection{Main Results}
As we can see from Table~\ref{tab:main_results}, NER-BERT is able to consistently outperform BERT across all target domains, with a 4.87\% averaged F1-score improvement in the \textit{Target Only} setting and a 4.34\% averaged F1-score improvement in the \textit{Source \& Target} setting. Moreover, we find that NER-BERT significantly surpasses BERT on the politics, science, music, literature and AI domains (all coming from the CrossNER dataset~\cite{liu2021crossner}), with a 5\% to 10\% F1-score improvement on each domain. We argue that it is difficult for BERT to achieve good performance on these domains since their data sizes for training are relatively small (only 100 or 200 training examples on these domains compared to several thousand examples for other domains) and there are abundant entity types that the model needs to categorize. In contrast, NER-BERT can better learn to recognize and categorize the entities on these domains since it has been pre-trained on a large-scale NER dataset with various entity categories. We observe that NER-BERT only marginally outperforms BERT in the Twitter domain (``Twi.'' and ``BTwi.''). This is because the training examples in ``Twi.'' and ``BTwi.'' are relatively large (around 5K), and the entity categories are limited (3 to 4 types), which makes it easier for BERT to capture the task information in this domain and narrow down the performance gap between BERT and NER-BERT.

In addition, NER-BERT also significantly outperforms CDLM and DAPT on the politics, science, music, literature and AI domains. Given that both DAPT and CDLM leverage an enormous unlabeled domain-related corpus to inject domain knowledge into pre-trained models in order to boost their domain adaptation ability, the further performance improvements from NER-BERT can be attributed to the relatively high quality of our constructed NER corpus, which makes our entity tagging-based pre-training more effective than the pre-training using the domain-related corpus.

Furthermore, we find that pre-training using coarse-grained entity types will greatly lower the effectiveness of the pre-training. From Table~\ref{tab:main_results}, we can see that although NER-BERT$_{\text{4types}}$ can outperform BERT in both the \textit{Target Only} and \textit{Source \& Target} settings, it consistently performs worse than NER-BERT, with a performance gap of more than 2\% averaged F1-score for both settings. This is because in the pre-training stage, it is difficult for NER-BERT$_{\text{4types}}$ to well learn the knowledge from various entities given the limited entity categories, which leads to a less effective pre-training. Therefore, with a larger entity category list, NER-BERT$_{\text{212types}}$ consistently outperforms NER-BERT$_{\text{4types}}$ across all target domains. Interestingly, we observe that NER-BERT (with 315 entity categories) can marginally outperform NER-BERT$_{\text{212types}}$ in most of the target domains and achieve slightly better performance on the averaged F1-score for both experimental settings. We conjecture that NER-BERT is pre-trained using a more abundant entity categories than NER-BERT$_{\text{212types}}$, and this helps NER-BERT extract more valuable entity-related knowledge from the constructed NER corpus, which assists in fast domain adaptation.

\subsection{Few-shot Settings}
To further study the effectiveness of NER-BERT in the extremely low-resource scenario where the number of training examples is around or less than 100.
Note that the number of training examples on the politics, science, music, literature, and AI domains is already small enough, so we conduct few-shot experiments using different percentages of the training data on the other six domains.
As illustrated in Figure~\ref{fig:few-shot}, we can see that NER-BERT is able to significantly improve the performance in all target domains (5\% to 10\% F1-score improvement), when the number of training examples is around or less than 50. This is because the task-specific pre-training injects the NER task-related knowledge into NER-BERT, which allows it to easily learn to categorize entities and quickly adapt to a new domain. By contrast, BERT greatly loses its effectiveness when only a few training examples are available since the large task discrepancy between the language modeling and NER task makes it difficult for BERT to quickly learn to categorize entities in a new domain.

\section{Visualization}
To further study the effectiveness of NER-BERT, we aim to visualize the entity representations across various entity categories for NER-BERT and the baselines.
To do so, we first input sentences into the pre-trained models and obtain their embeddings for the first token of each entity. Then we reduce the high-dimension embeddings to a two-dimensional point using tSNE.
As shown in Figure~\ref{fig:visualization}, we visualize the entity representations for BERT, NER-BERT$_{\text{4types}}$, NER-BERT$_{\text{212types}}$, and NER-BERT in the test set of the politics domain.~\footnote{Visualizations for different domains are in Appendix~\ref{Appendix_C}.}
We observe that it is hard to find clear group boundaries for almost all entity categories for the BERT model. Entity representations for NER-BERT$_{\text{4types}}$ has relatively clearer group boundaries compared to those for BERT, while the model still cannot well distinguish the fine-grained categories (e.g., ``party'' (orange) and ``organization'' (green)) since it is pre-trained using only four entity types.
Interestingly, we find that the boundaries of entity categories are generally clear, except for the ``miscellaneous'' category (black), for both NER-BERT$_{\text{212types}}$ and NER-BERT. This is because both models are pre-trained using numerous entity categories, which provides them the good pre-learned knowledge for categorizing a variety of entities.

\section{Related Work}
\subsection{Low-Resource NER}
Low-resource NER models aim to enhance the model's performance on the entity tagging task when only a few annotated data are available~\cite{ghaddar2018transforming,cao2019low,liu2020coach,jia2019cross,liu2020zero,liu2021importance,jia2020multi,liu2021crossner}.
\citet{jia2019cross} proposed a cross-domain language modeling approach to boost the domain adaptation performance in the NER task. \citet{liu2021crossner} collected five diverse domains for the NER task, and incorporated domain-adaptive pre-training and task-adaptive pre-training~\cite{gururangan2020don} that use unlabeled domain-related corpora to improve the NER performance in low-resource scenarios. 
Instead of leveraging unlabeled data, \citet{cao2019low} generated weakly-labeled NER data based on Wikipedia anchors and a taxonomy for low-resource languages. However, the entity categories for the created data are limited to a few coarse-grained types. \citet{mengge2020coarse} used k-means clustering to extract fine-grained categories from the coarse-grained types. Nevertheless, obtaining fine-grained entities based on clustering is not stable, and we have to manually select the number of clusters based on the constructed dataset.

\subsection{Pre-trained Models}
Recently, pre-training has become an indispensable part of developing algorithms and building models for almost all natural language processing tasks~\cite{devlin2019bert,liu2019roberta,yamada2020luke,wu2020tod,liu2021crossner}.
Pre-trained language models that are trained on large-scale plain text corpora such as Wikipedia and BookCorpus~\cite{zhu2015aligning} have achieved promising results in natural language understanding and generation tasks~\cite{peters2018deep,devlin2019bert,yang2019xlnet,liu2019roberta,lewis2020bart,raffel2020exploring}, and have been shown to be effective in a data scarcity scenario~\cite{ma2019domain,radford2019language,chen2020few}. Due to the underlying discrepancies between the language modeling and downstream tasks, task-specific pre-training methods have been proposed to further boost the task performance, such as SciBERT~\cite{beltagy2019scibert}, VideoBERT~\cite{sun2019videobert}, DialoGPT~\cite{zhang2020dialogpt}, PLATO~\cite{bao2020plato}, CodeBERT~\cite{feng2020codebert}, ToD-BERT~\cite{wu2020tod} and VL-BERT~\cite{Su2020VLBERT}.

\section{Conclusion}
In this paper, we first incorporate Wikipedia anchors and DBpedia Ontology to build a large-scale NER dataset with a relatively high quality. Then, we utilize the constructed dataset to pre-train NER-BERT.
Results illustrate that it is essential to leverage various entity categories for pre-training, and NER-BERT is able to significantly outperform BERT as well as other strong baselines across nine diverse domains. Additionally, we show that NER-BERT is especially effective when only a few pre-training examples are available in target domains. Moreover, the visualization further indicates that NER-BERT possesses good pre-learned knowledge for categorizing a variety of entities.

\bibliography{custom}
\bibliographystyle{acl_natbib}

\clearpage
\appendix

\section{Data Statistics}
\label{Appendix_A}
\begin{table*}[]
\centering
\begin{adjustbox}{width={0.83\textwidth},totalheight={\textheight},keepaspectratio}
\begin{tabular}{l|ccccccccccc}
\toprule
      & \textbf{Pol.} & \textbf{Sci.} & \textbf{Mus.} & \textbf{Lit.} & \textbf{AI}  & \textbf{Twi.} & \textbf{BTwi.} & \textbf{BioCG} & \textbf{BioPC} & \textbf{Fin.} & \textbf{Def.} \\ \midrule
\textbf{Train} & 200  & 200  & 100  & 100  & 100 & 4.3K & 6.3K  & 3.0K  & 2.5K  & 992  & 611  \\
\textbf{Dev}   & 541  & 450  & 380  & 400  & 350 & 1.4K & 1.0K  & 1.0K  & 0.9K  & 176  & 153  \\
\textbf{Test}  & 651  & 543  & 456  & 416  & 431 & 1.5K & 2.0K  & 1.9K  & 1.7K  & 305  & 199 \\
\bottomrule
\end{tabular}
\end{adjustbox}
\caption{Data statistics of train, dev and test sets for all domains.}
\label{tab:statistics}
\end{table*}

The data statistics for all the domains are shown in Table~\ref{tab:statistics}.

\section{Entity Categories}
\label{Appendix_B}
The 315 entity categories are listed as follows: ``populatedplace'', ``asteroid'', ``amphibian'', ``vein'', ``garden'', ``judge'', ``cheese'', ``horserider'', ``baseballseason'', ``artery'', ``fashion'', ``comicstrip'', ``moss'', ``poem'', ``poet'', ``monoclonalantibody'', ``cricketground'', ``archaea'', ``voiceactor'', ``nerve'', ``classicalmusiccomposition'', ``beachvolleyballplayer'', ``photographer'', ``cyclingteam'', ``reptile'', ``educationalinstitution'', ``entomologist'', ``lacrosseplayer'', ``bodybuilder'', ``sportsteammember'', ``ambassador'', ``artistdiscography'', ``golfcourse'', ``businessperson'', ``muscle'', ``rollercoaster'', ``brewery'', ``formermunicipality'', ``handballteam'', ``winery'', ``hollywoodcartoon'', ``mammal'', ``netballplayer'', ``volleyballleague'', ``crater'', ``mythologicalfigure'', ``arachnid'', ``squashplayer'', ``roadjunction'', ``colour'', ``musicfestival'', ``roadtunnel'', ``televisionseason'', ``railwaystation'', ``college'', ``eukaryote'', ``lawfirm'', ``priest'', ``bone'', ``cave'', ``stadium'', ``brain'', ``biologicaldatabase'', ``historian'', ``screenwriter'', ``cultivatedvariety'', ``animangacharacter'', ``tabletennisplayer'', ``restaurant'', ``railwaytunnel'', ``glacier'', ``amusementparkattraction'', ``volleyballplayer'', ``horsetrainer'', ``wineregion'', ``handballplayer'', ``skater'', ``galaxy'', ``pokerplayer'', ``medician'', ``mineral'', ``speedwayrider'', ``monument'', ``crustacean'', ``siteofspecialscientificinterest'', ``bacteria'', ``soccerclubseason'', ``writer'', ``skiarea'', ``species'', ``radiohost'', ``painter'', ``device'', ``anime'', ``journalist'', ``mayor'', ``memberofparliament'', ``archbishop'', ``enzyme'', ``motorcycle'', ``jockey'', ``automobileengine'', ``fish'', ``wrestlingevent'', ``lighthouse'', ``senator'', ``chef'', ``politician'', ``badmintonplayer'', ``astronaut'', ``animal'', ``mollusca'', ``beautyqueen'', ``skier'', ``hotel'', ``rocket'', ``outbreak'', ``dartsplayer'', ``gymnast'', ``fungus'', ``curler'', ``volcano'', ``castle'', ``engineer'', ``mixedmartialartsevent'', ``powerstation'', ``beverage'', ``motorcyclerider'', ``mountainpass'', ``protein'', ``congressman'', ``prison'', ``grape'', ``manga'', ``cyclingrace'', ``fashiondesigner'', ``star'', ``programminglanguage'', ``anatomicalstructure'', ``model'', ``gaelicgamesplayer'', ``sportsleague'', ``artwork'', ``swimmer'', ``combinationdrug'', ``earthquake'', ``olympicevent'', ``drug'', ``train'', ``scientist'', ``shoppingmall'', ``insect'', ``cardinal'', ``economist'', ``christianbishop'', ``musical'', ``figureskater'', ``plant'', ``event'', ``radioprogram'', ``filmfestival'', ``tennisplayer'', ``food'', ``supremecourtoftheunitedstatescase'', ``golftournament'', ``chessplayer'', ``locomotive'', ``governor'', ``lake'', ``canal'', ``grandprix'', ``motorsportseason'', ``actor'', ``planet'', ``humangene'', ``tradeunion'', ``racecourse'', ``primeminister'', ``musicalartist'', ``tennistournament'', ``bridge'', ``worldheritagesite'', ``play'', ``dam'', ``noble'', ``buscompany'', ``televisionepisode'', ``basketballleague'', ``library'', ``footballmatch'', ``cyclist'', ``bank'', ``location'', ``religiousbuilding'', ``park'', ``criminal'', ``academicjournal'', ``ship'', ``spacemission'', ``currency'', ``comic'', ``historicbuilding'', ``boxer'', ``airline'', ``chemicalcompound'', ``theatre'', ``hospital'', ``sportsteam'', ``cleric'', ``holiday'', ``golfplayer'', ``icehockeyleague'', ``sportsevent'', ``formulaoneracer'', ``rugbyleague'', ``comedian'', ``martialartist'', ``architect'', ``footballteam'', ``president'', ``radiostation'', ``diocese'', ``ncaateamseason'', ``horserace'', ``comicscreator'', ``informationappliance'', ``pope'', ``collegecoach'', ``baseballleague'', ``musicgenre'', ``soapcharacter'', ``weapon'', ``rugbyclub'', ``rugbyplayer'', ``convention'', ``disease'', ``icehockeyplayer'', ``militarystructure'', ``publisher'', ``cricketteam'', ``videogame'', ``saint'', ``comicscharacter'', ``artist'', ``protectedarea'', ``broadcastnetwork'', ``railwayline'', ``athlete'', ``airport'', ``historicplace'', ``cricketer'', ``aircraft'', ``automobile'', ``basketballteam'', ``racingdriver'', ``philosopher'', ``basketballplayer'', ``soccertournament'', ``citydistrict'', ``recordlabel'', ``hockeyteam'', ``wrestler'', ``software'', ``song'', ``bodyofwater'', ``village'', ``footballleagueseason'', ``publictransitsystem'', ``sport'', ``station'', ``mountain'', ``televisionstation'', ``soccermanager'', ``magazine'', ``museum'', ``governmentagency'', ``film'', ``venue'', ``baseballplayer'', ``writtenwork'', ``fictionalcharacter'', ``album'', ``website'', ``award'', ``school'', ``building'', ``island'', ``footballplayer'', ``election'', ``militaryperson'', ``soccerplayer'', ``soccerleague'', ``politicalparty'', ``newspaper'', ``river'', ``legislature'', ``road'', ``ethnicgroup'', ``televisionshow'', ``language'', ``university'', ``band'', ``town'', ``royalty'', ``militaryunit'', ``organisation'', ``militaryconflict'', ``company'', ``soccerclub'', ``administrativeregion'', ``city'', ``settlement'', ``person'', ``country'', ``ENTITY''.

\begin{figure*}[!ht]
\centering
\begin{subfigure}{.49\textwidth}
  \centering
  \includegraphics[width=0.85\linewidth]{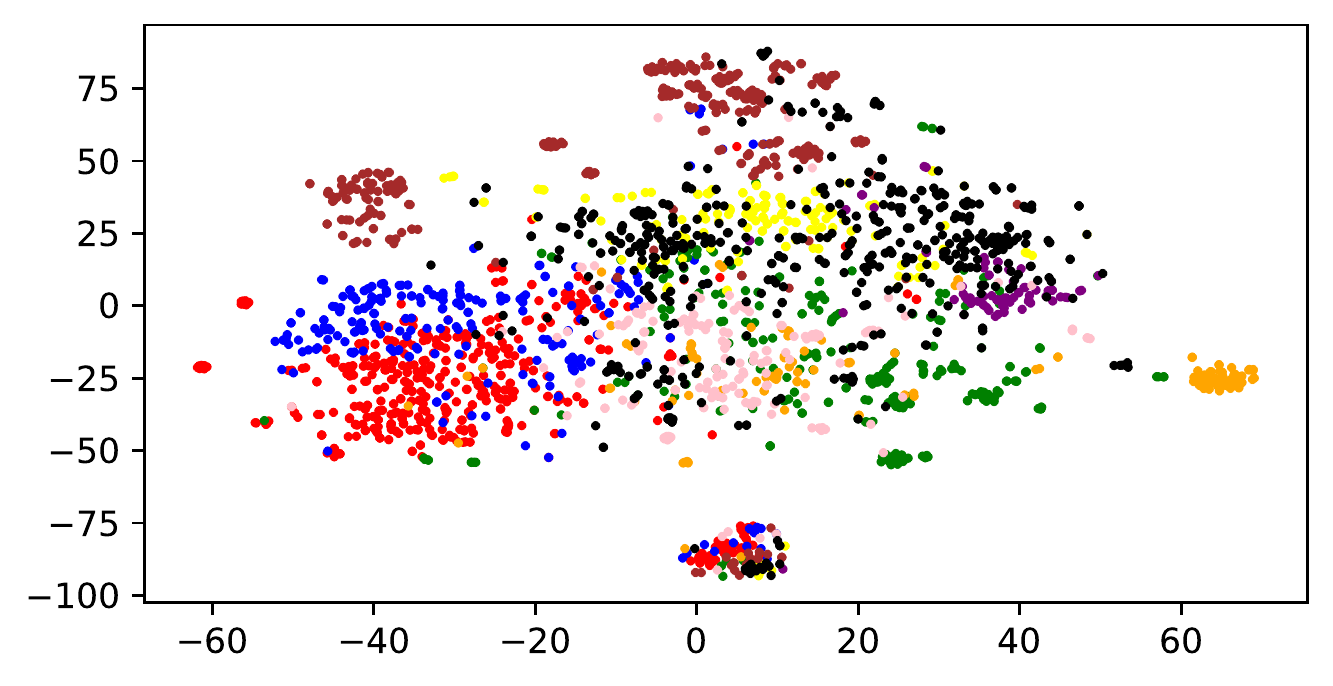}  
  \caption{BERT}
\end{subfigure}
\begin{subfigure}{.49\textwidth}
  \centering
  \includegraphics[width=0.85\linewidth]{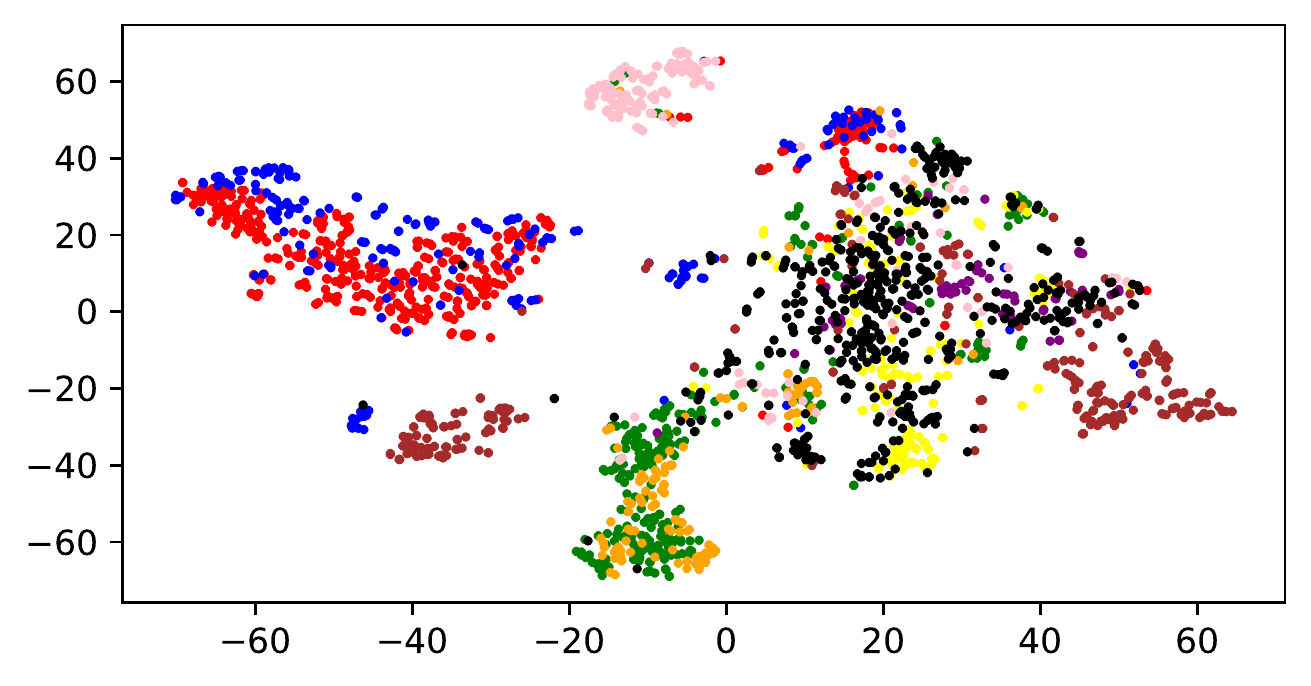}
  \caption{NER-BERT$_{\text{4types}}$}
\end{subfigure}
\begin{subfigure}{.49\textwidth}
  \centering
  \includegraphics[width=0.85\linewidth]{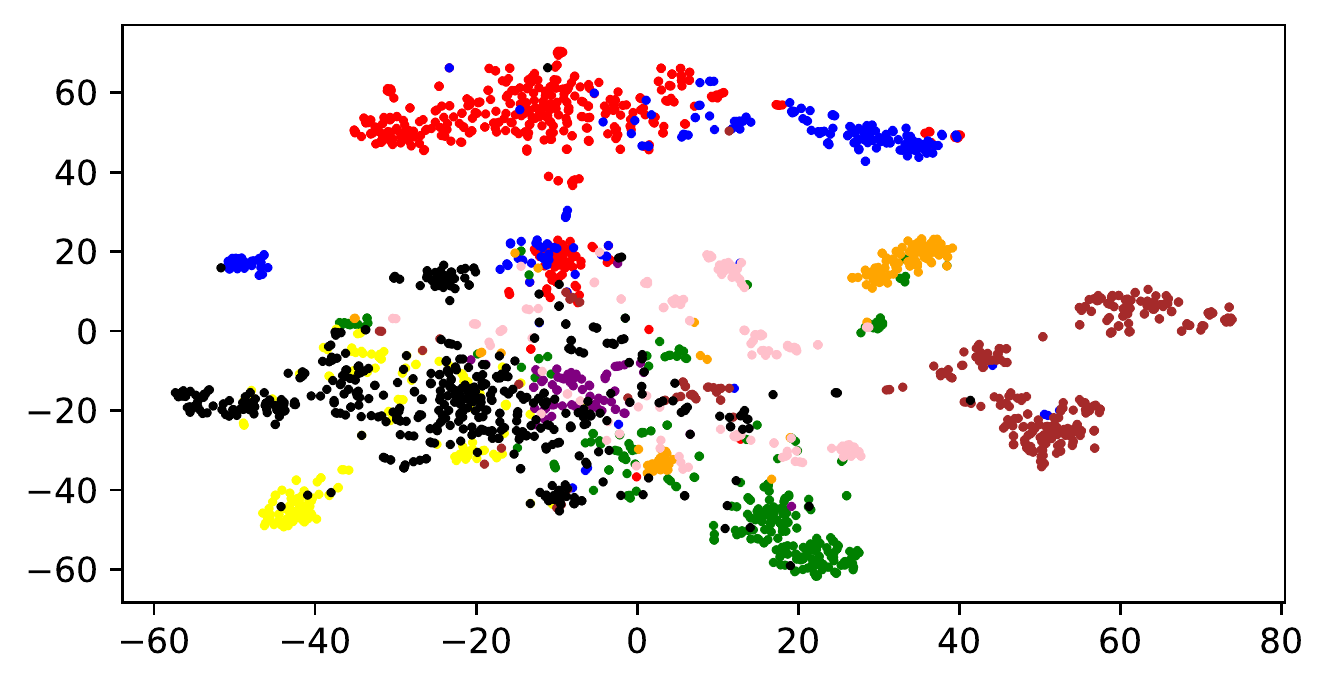}
  \caption{NER-BERT$_{\text{212types}}$}
\end{subfigure}
\begin{subfigure}{.49\textwidth}
  \centering
  \includegraphics[width=0.85\linewidth]{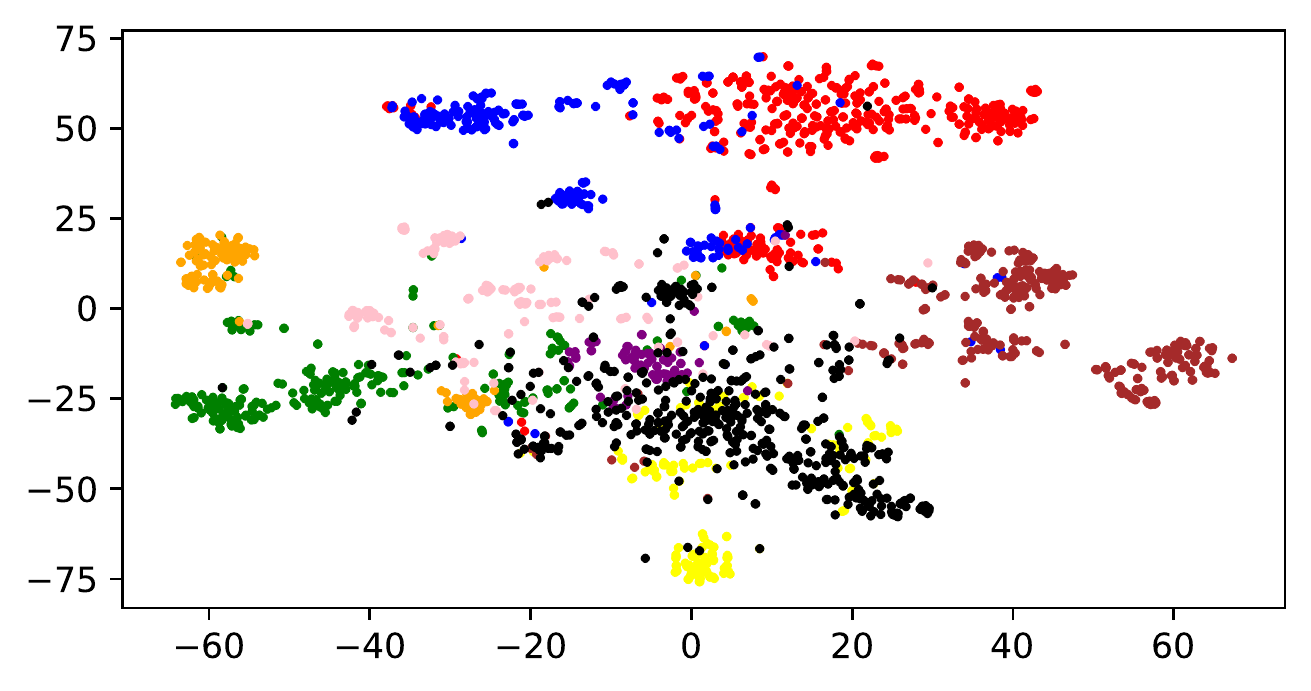}
  \caption{NER-BERT}
\end{subfigure}
\caption{The tSNE visualization of entity representations in the test set of the \textbf{science} domain.}
\label{fig:visualization_science}
\end{figure*}

\begin{figure*}[!ht]
\centering
\begin{subfigure}{.49\textwidth}
  \centering
  \includegraphics[width=0.85\linewidth]{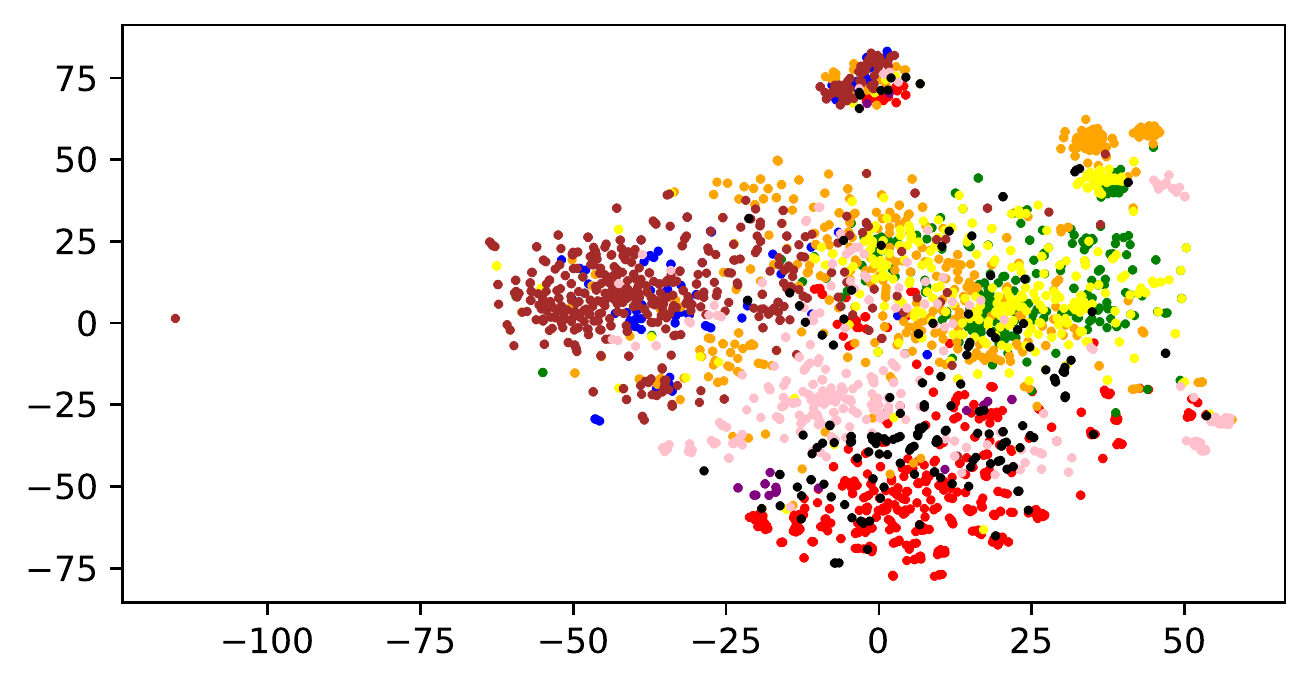}  
  \caption{BERT}
\end{subfigure}
\begin{subfigure}{.49\textwidth}
  \centering
  \includegraphics[width=0.85\linewidth]{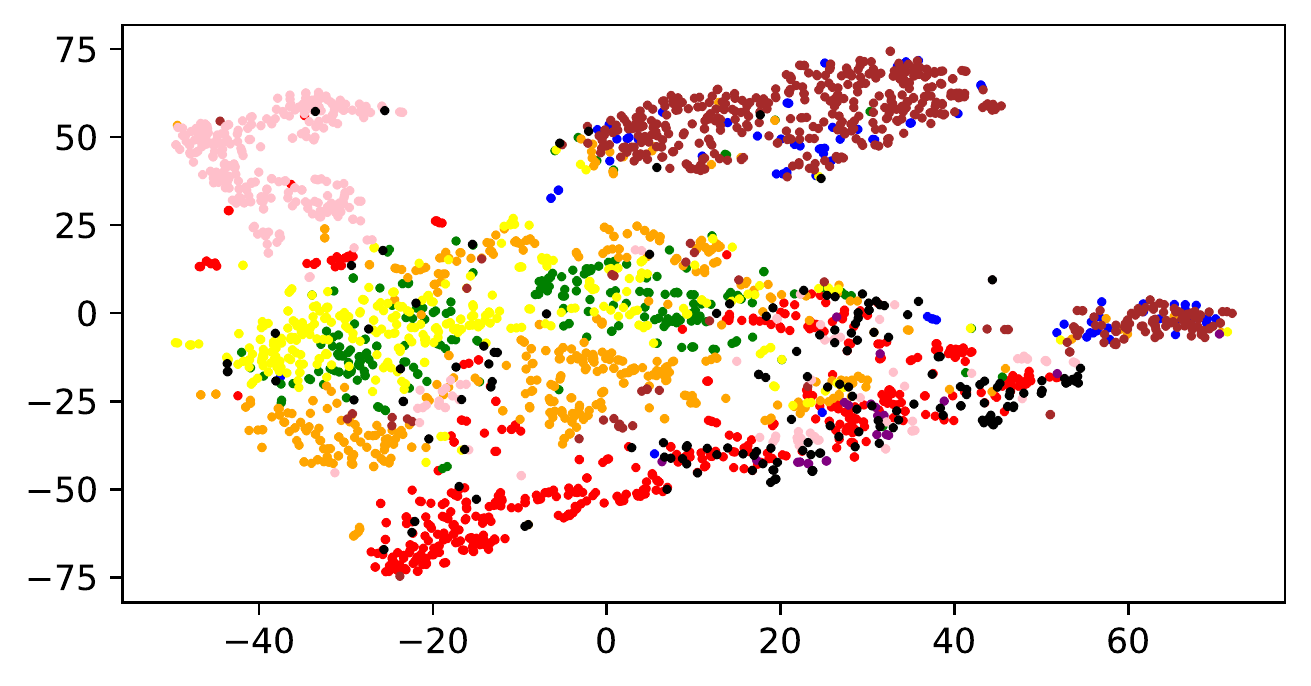}
  \caption{NER-BERT$_{\text{4types}}$}
\end{subfigure}
\begin{subfigure}{.49\textwidth}
  \centering
  \includegraphics[width=0.85\linewidth]{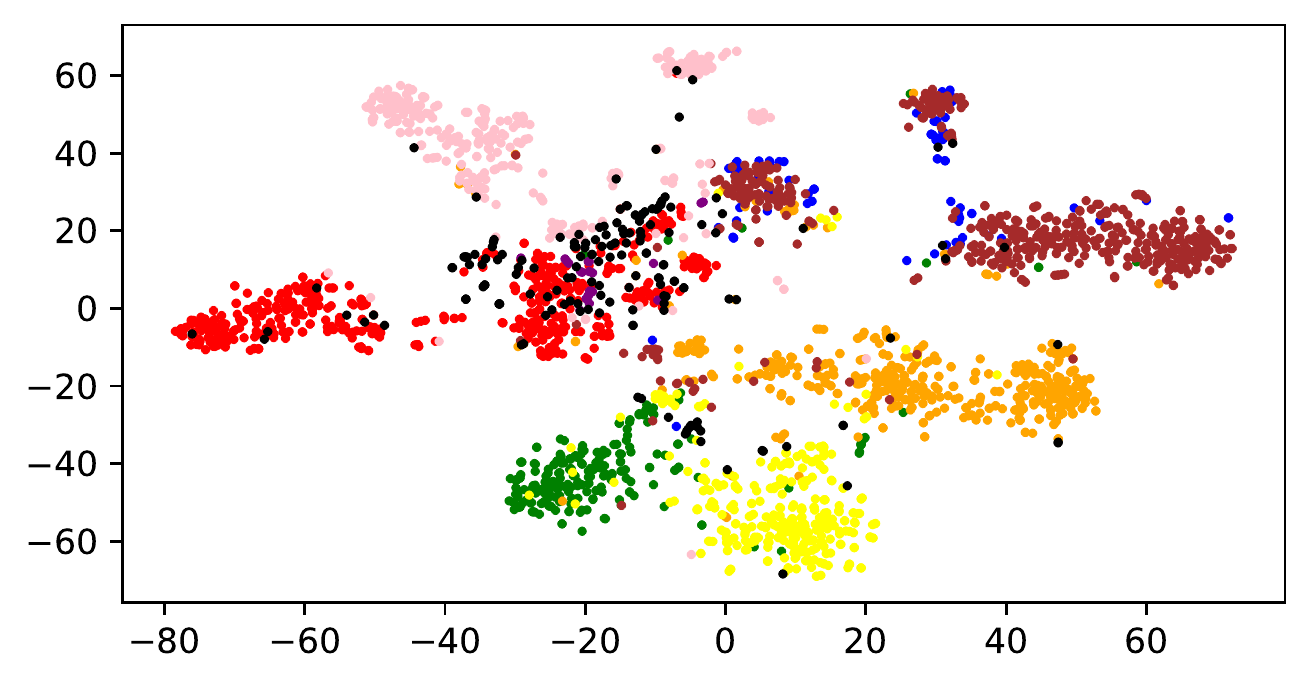}
  \caption{NER-BERT$_{\text{212types}}$}
\end{subfigure}
\begin{subfigure}{.49\textwidth}
  \centering
  \includegraphics[width=0.85\linewidth]{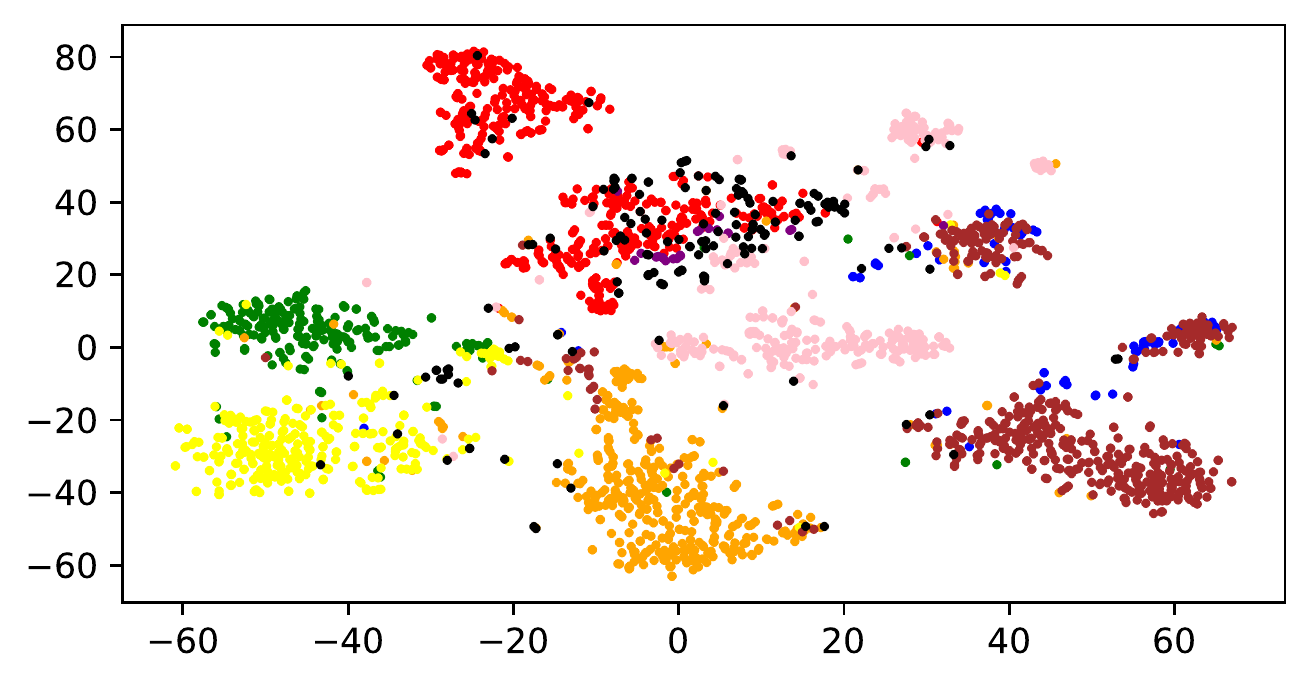}
  \caption{NER-BERT}
\end{subfigure}
\caption{The tSNE visualization of entity representations in the test set of the \textbf{music} domain.}
\label{fig:visualization_music}
\end{figure*}

\begin{figure*}[!ht]
\centering
\begin{subfigure}{.49\textwidth}
  \centering
  \includegraphics[width=0.85\linewidth]{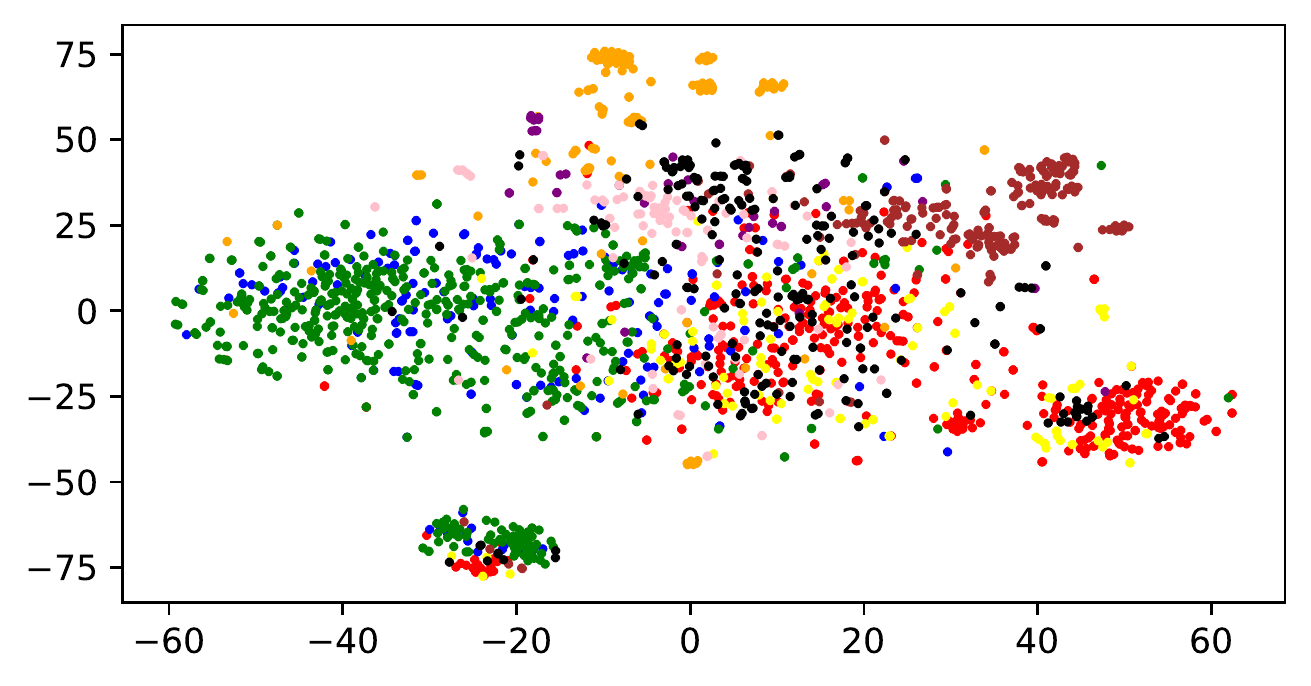}  
  \caption{BERT}
\end{subfigure}
\begin{subfigure}{.49\textwidth}
  \centering
  \includegraphics[width=0.85\linewidth]{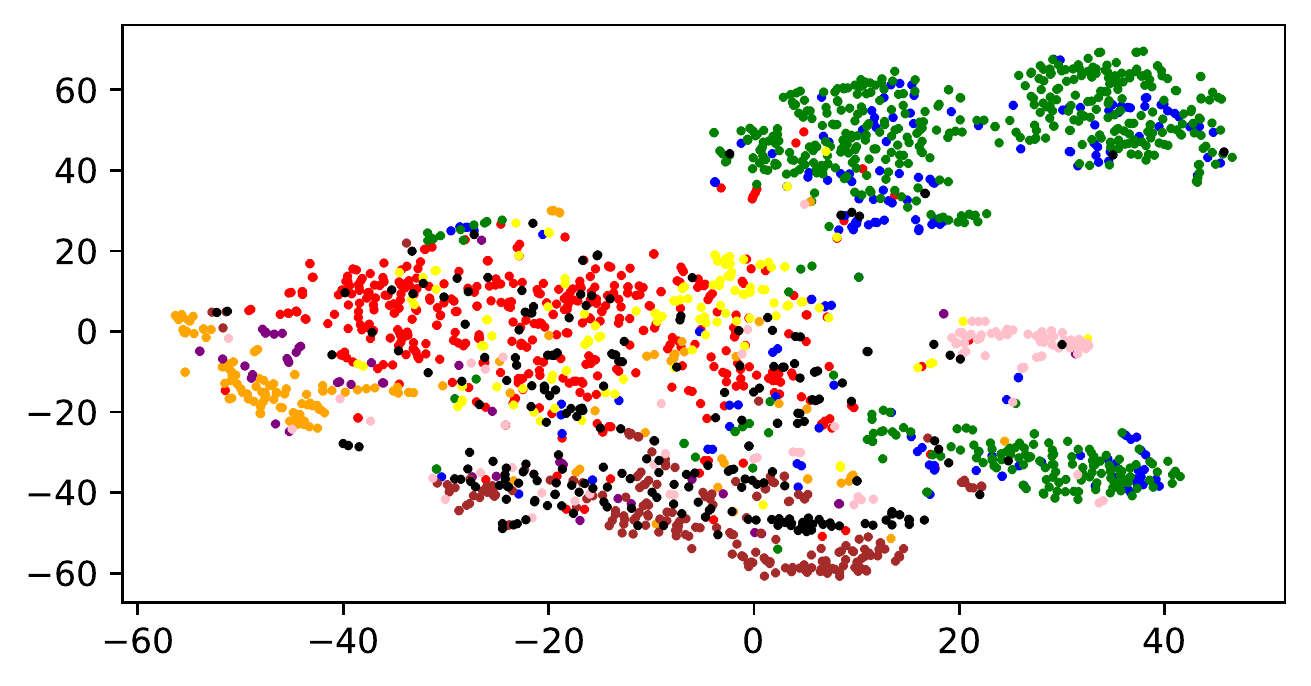}
  \caption{NER-BERT$_{\text{4types}}$}
\end{subfigure}
\begin{subfigure}{.49\textwidth}
  \centering
  \includegraphics[width=0.85\linewidth]{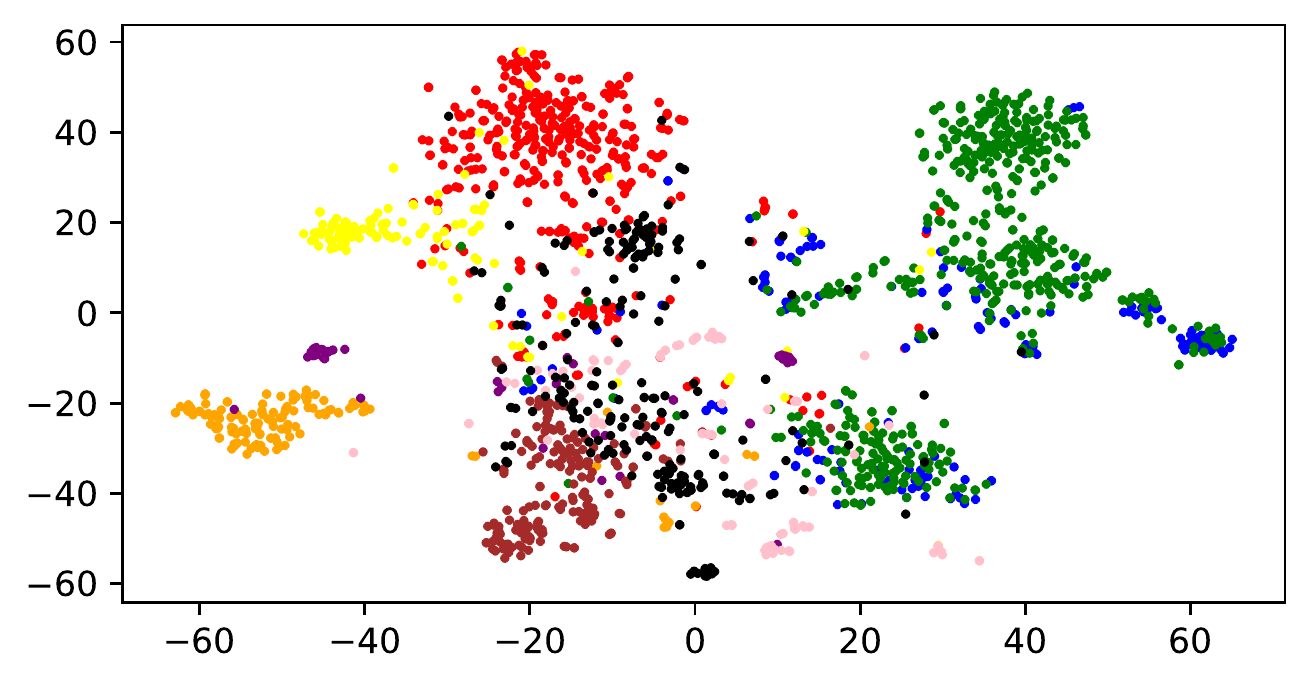}
  \caption{NER-BERT$_{\text{212types}}$}
\end{subfigure}
\begin{subfigure}{.49\textwidth}
  \centering
  \includegraphics[width=0.85\linewidth]{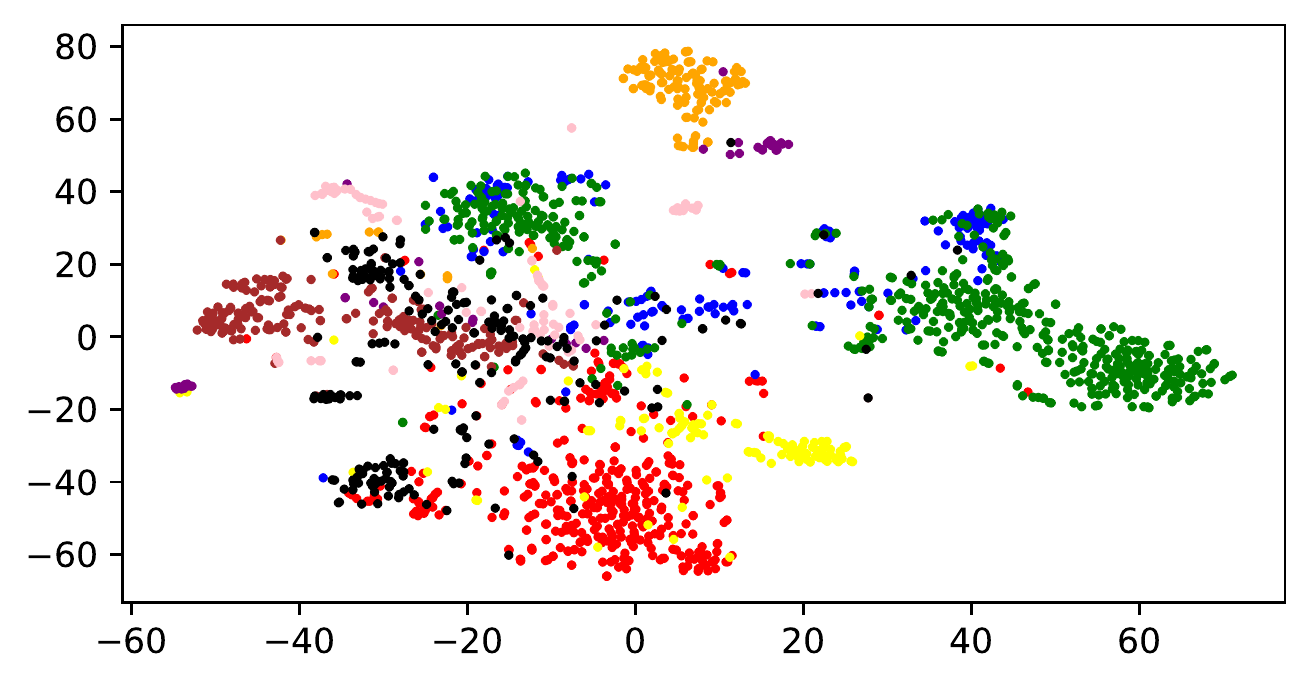}
  \caption{NER-BERT}
\end{subfigure}
\caption{The tSNE visualization of entity representations in the test set of the \textbf{literature} domain.}
\label{fig:visualization_literature}
\end{figure*}

\begin{figure*}[!ht]
\centering
\begin{subfigure}{.49\textwidth}
  \centering
  \includegraphics[width=0.85\linewidth]{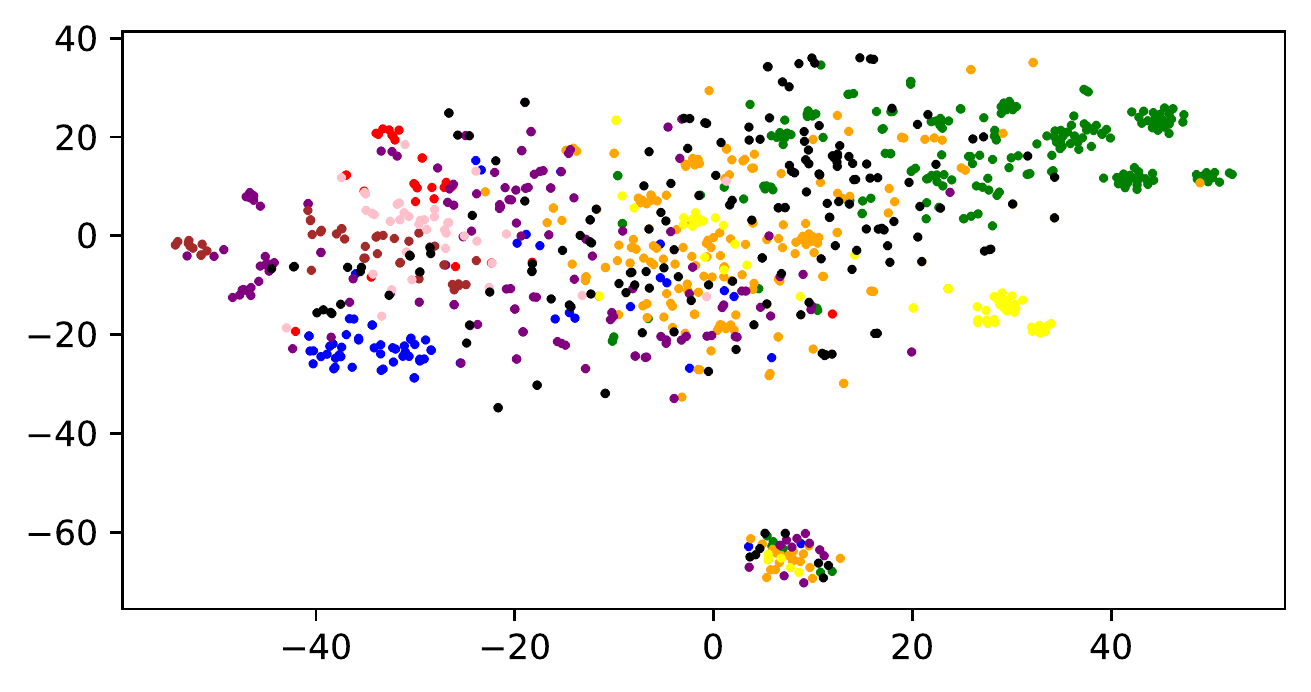}  
  \caption{BERT}
\end{subfigure}
\begin{subfigure}{.49\textwidth}
  \centering
  \includegraphics[width=0.85\linewidth]{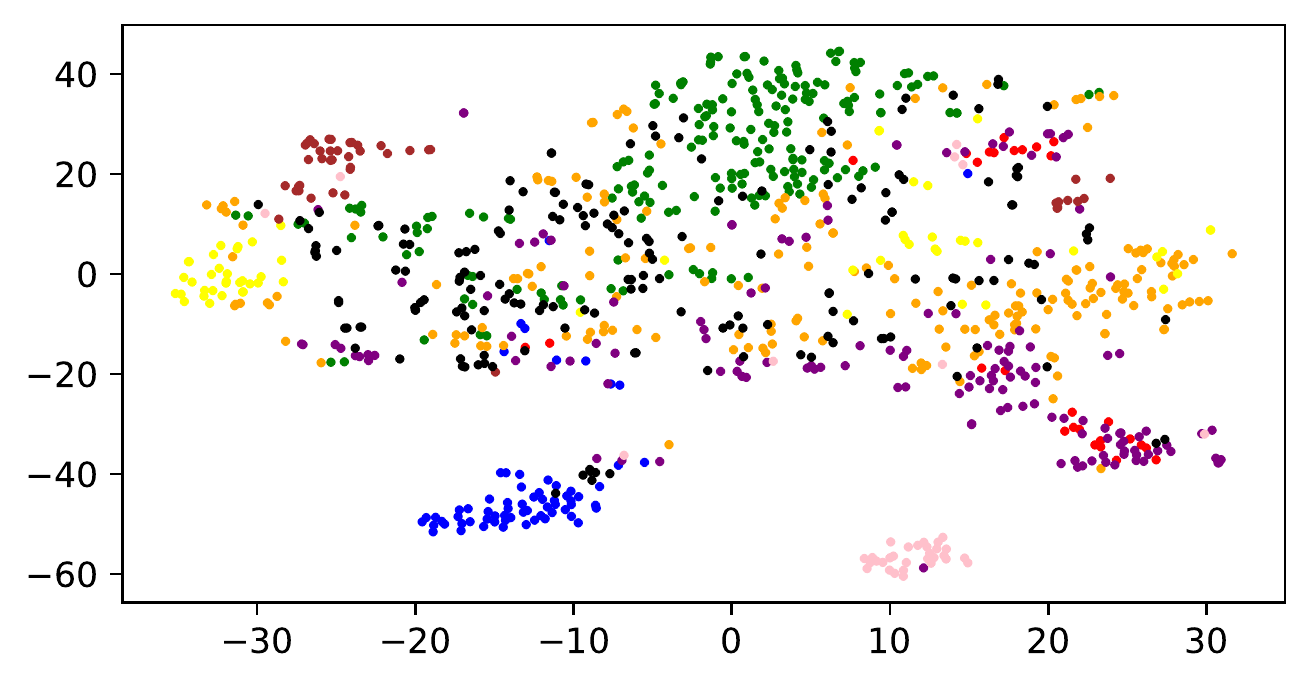}
  \caption{NER-BERT$_{\text{4types}}$}
\end{subfigure}
\begin{subfigure}{.49\textwidth}
  \centering
  \includegraphics[width=0.85\linewidth]{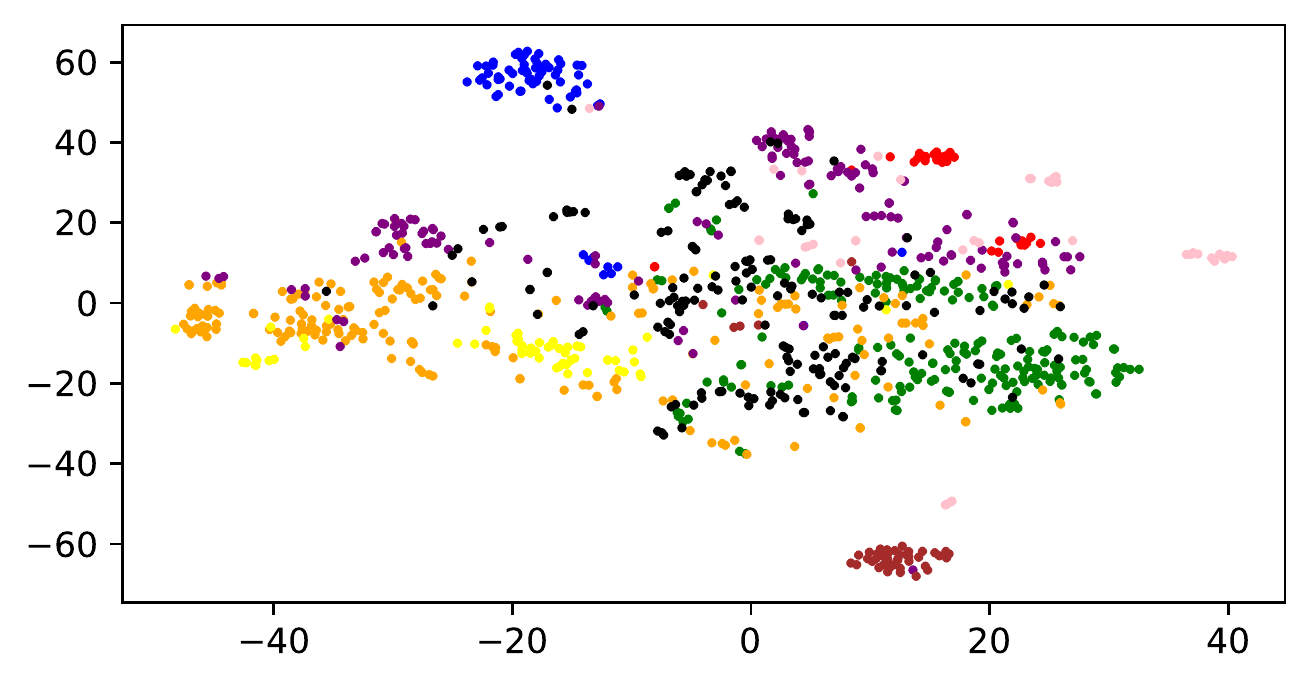}
  \caption{NER-BERT$_{\text{212types}}$}
\end{subfigure}
\begin{subfigure}{.49\textwidth}
  \centering
  \includegraphics[width=0.85\linewidth]{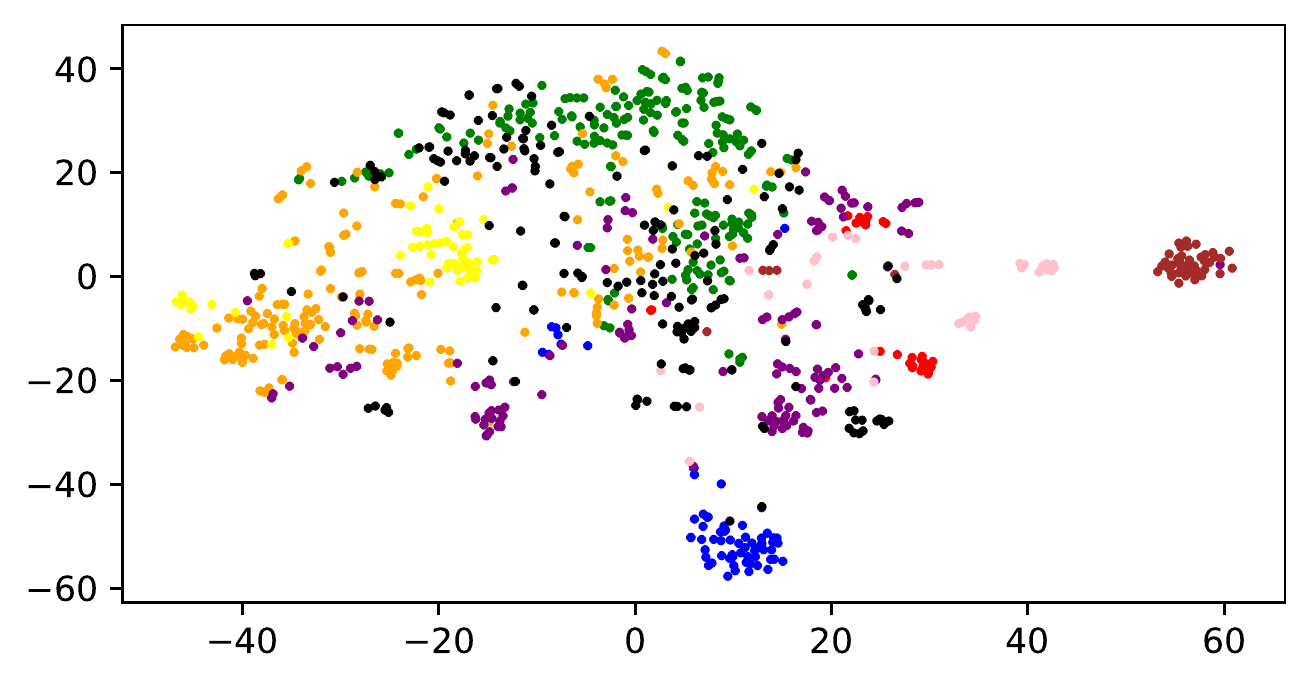}
  \caption{NER-BERT}
\end{subfigure}
\caption{The tSNE visualization of entity representations in the test set of the \textbf{AI} domain.}
\label{fig:visualization_ai}
\end{figure*}

\section{Visualization}
\label{Appendix_C}
The visualization of entity representations in the test set of science, music, literature and AI domains are shown in Figure~\ref{fig:visualization_science}, Figure~\ref{fig:visualization_music}, Figure~\ref{fig:visualization_literature}, and Figure~\ref{fig:visualization_ai}.

\begin{figure*}[!ht]
    \centering
    \resizebox{0.29\textwidth}{!}{  \includegraphics{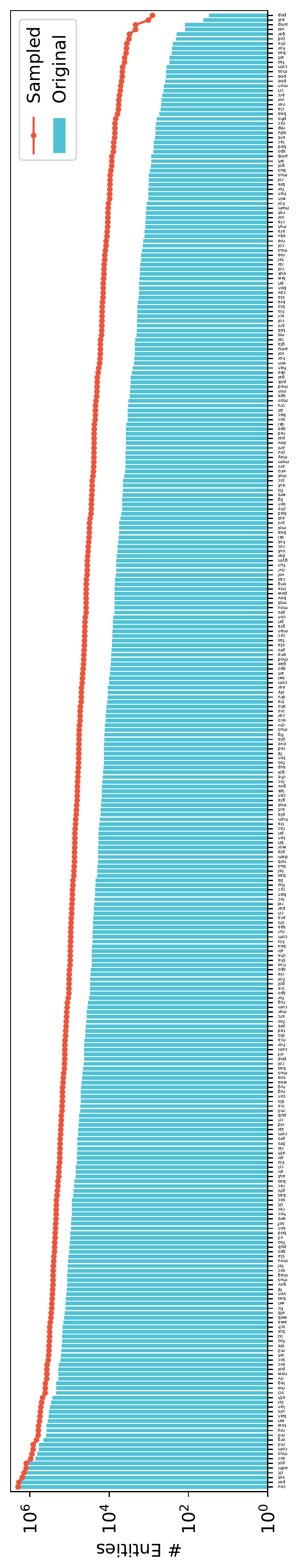}
    }
    \caption{Number of entities before and after the sampling. X-axis is a list of categories that are denoted by their first three characters.}
    \label{fig:category_full}
\end{figure*}

\section{Entity Category Ratio}
\label{Appendix_D}
The number of entities before and after the sampling (full table) is illustrated in Figure~\ref{fig:category_full}


\end{document}